\documentclass{article}

\usepackage{PRIMEarxiv}

\usepackage[utf8]{inputenc} 
\usepackage[T1]{fontenc}    
\usepackage{hyperref}       
\usepackage{url}            
\usepackage{booktabs}       
\usepackage{amsfonts}       
\usepackage{nicefrac}       
\usepackage{microtype}      
\usepackage{lipsum}
\usepackage{fancyhdr}       
\usepackage{graphicx}       
\usepackage{amsmath}
\usepackage{subcaption}
\usepackage{placeins}
\usepackage{multirow}
\usepackage{caption}
\usepackage{subcaption}
\graphicspath{{media/}}     

\usepackage{floatrow}
\newfloatcommand{capbtabbox}{table}[][\FBwidth]

\newcommand{\KL}{\mathrm{KL}}

\title{Efficient-VDVAE: Less is more
}

\author{
  Louay Hazami\thanks{Equal contribution.} \ \thanks{To whom correspondence should be addressed.} \\
  Cash App Labs \\
  \texttt{louay@squareup.com} \\
  \And
  Rayhane Mama\footnotemark[1] \\
  Cash App Labs \\
  \texttt{rayhane@squareup.com} \\
  \And
  Ragavan Thurairatnam \\
  Cash App Labs \\
  \texttt{ragavan@squareup.com} \\
}

\begin{document}
\maketitle

\begin{abstract}
Hierarchical VAEs have emerged in recent years as a reliable option for maximum likelihood estimation. However, instability issues and demanding computational requirements have hindered research progress in the area. We present simple modifications to the Very Deep VAE to make it converge up to $2.6\times$ faster, save up to $20\times$ in memory load and improve stability during training. Despite these changes, our models achieve comparable or better negative log-likelihood performance than current state-of-the-art models on all $7$ commonly used image datasets we evaluated on. We also make an argument against using 5-bit benchmarks as a way to measure hierarchical VAE's performance due to undesirable biases caused by the 5-bit quantization. Additionally, we empirically demonstrate that roughly $3\%$ of the hierarchical VAE's latent space dimensions is sufficient to encode most of the image information, without loss of performance, opening up the doors to efficiently leverage the hierarchical VAEs' latent space in downstream tasks. We release our source code and models at \href{https://github.com/Rayhane-mamah/Efficient-VDVAE}{\url{https://github.com/Rayhane-mamah/Efficient-VDVAE}}.
\end{abstract}


\section{Introduction}
Maximum likelihood models have garnered a lot of attention from researchers in the last few years. Deep autoregressive models\cite{van2016conditional, salimans2017pixelcnn++, sadeghi2019pixelvae++, augmented_sparse_transformer} have long achieved the best log likelihoods across different modalities. Variational Autoencoder\cite{kingma2014autoencoding, rezende2014stochastic} (VAE) is a different type of maximum likelihood models that are usually associated with representation learning \cite{bengio_rl, radford_rl} and are known to generate subjectively more diverse images\cite{vae_diversity} compared to other generative approaches like diffusion models \cite{diffusion_1, diffusion_2, diffusion_3} and GANs \cite{radford_rl, gan1, gan2, gan3}. Very Deep VAE (VDVAE) \cite{child2021very}, a newly proposed hierarchical VAE (HVAE), shows great promise in competing with deep autoregressive models in Maximum Likelihood Estimation (MLE). However, this method has proven to be computationally costly and to be generally unstable during training. In this paper, we aim to tackle these two issues and to share some of our insights that we have gathered throughout our experiments with VDVAE.

We start this work with a background on VAEs and HVAEs in section \ref{background} and list similar works to VDVAE in section \ref{related_work}. Both these sections can be skipped by practitioners who are familiar with HVAEs.

In section \ref{contributions}, we introduce Efficient-VDVAE, which encompasses our contributions and tackles both the instability and compute cost problems associated with VDVAE.

We kick off the section \ref{experiments} by empirically showing the impact of our approach. We then provide evaluations of our model trained on 7 commonly used benchmarks and show that our models are either on par with VDVAE or better in terms of Negative Log Likelihood (NLL) on all datasets. We also share some generated samples from the prior distribution of our model.

We then interpret the Efficient-VDVAE from an informational theory perspective in section \ref{lossless_compression}, and we show the dilemma between that interpretation and the current dominant HVAE implementations, demonstrating along the way that commonly used 5-bit benchmarks are not suitable benchmarks for HVAEs.

Finally, in section \ref{effective_latent_space}, we examine the size of the "effective latent space" of the Efficient-VDVAE and illustrate that one can prune $97\%$ of the latent space without harming reconstruction quality, opening up the doors for a more effective usage of the HVAE's latent space in downstream tasks.

For practitioners interested in using VDVAE based models, we provide tips and insights gained from this work in the supplemental material \ref{tips_insights}.

\section{Background}
\label{background}
\subsection{Variational AutoEncoders}
The goal of VAEs is to learn a generative model parametrized with $\theta$ of a joint distribution $p_{\theta}(x, z)$ that is usually factorized as:

\begin{equation}
    p_\theta(x, z) = p_\theta(x | z)p_\theta(z)
\end{equation}

where $p_\theta(z)$ is a prior distribution over latent variables $z$ and $p_\theta(x | z)$ is the likelihood function (usually called stochastic decoder) that generates a data sample $x$ from its latent variables $z$. VAEs are trained with the Evidence Lower Bound (ELBO):

\begin{equation}\label{elbo_eq}
    \log p_\theta(x) \geq \mathop{\mathbb{E}}_{q_{\phi}(z|x)} \left[\log p_{\theta}(x|z)\right] - D_\KL\left[ q_\phi(z|x) || p_\theta(z)) \right]
\end{equation}

Where $D_\KL$ is the Kullback–Leibler divergence and $q_\phi(z|x)$ is the approximate posterior parametrized with $\phi$. Usually in typical VAEs, $q_\phi(z|x)$ and $p_\theta(z)$ are modeled with Gaussian distributions. For a derivation of the ELBO and a more in depth introduction to VAEs, see \cite{kingma2019vae_tutorial}.

\subsection{Hierarchical Variational AutoEncoders}

While fully factorized Gaussian $q_\phi(z|x)$ and $p_\theta(z)$ VAEs are theoretically capable of modeling any complex data distribution $p(x)$, in practice they are usually unable to learn such a solution due to computational constraints and a difficult optimization landscape\cite{yu2020frombayestoit}. To help alleviate this problem, deep HVAEs \cite{ranganath2016hierarchical, NIPS2016_ddeebdee, sonderby2016ladder, klushyn2019learning} have been proposed in recent years as a way to increase the expressiveness of both the approximate posterior and the prior distributions, such as:

\begin{equation}
\begin{split}
    p_\theta(z) = p_\theta(z_1) & \prod_{i=2}^{L} p_\theta(z_i | z_{<i}) \\
    q_\phi(z|x) = q_\phi(z_1 | x) & \prod_{i=2}^{L} q_\phi(z_i|x,z_{<i})
\end{split}
\end{equation}

\begin{equation} \label{generative_model_eq}
    p_\theta(x|z) = p_\theta(x|z_L) p_\theta(z) = p_\theta(x|z_L) p_\theta(z_1) \prod_{i=2}^{L} p_\theta(z_i | z_{<i})
\end{equation}

where $L$ is the total number of latent hierarchical variables $z = \{z_1, z_2, ..., z_L\}$. This formulation is known as the \textit{top-down inference} (or \textit{bidirectional inference}) HVAE introduced in \cite{sonderby2016ladder}. Figure \ref{bidirectional_hvae_fig} gives a general overview of this process.

By substituting the hierarchical stochastic encoder, stochastic decoder and prior expressions in equation (\ref{elbo_eq}), we obtain the Hierarchical ELBO that HVAEs are trained to maximize:

\begin{equation} \label{helbo_eq}
    \log p_\theta(x) \geq \mathop{\mathbb{E}}_{q_{\phi}(z|x)} \left[\log p_{\theta}(x|z)\right] - D_\KL\left[ q_{\phi}(z_1|x) || p_\theta(z_1)) \right] - \sum_{i=2}^{L} \mathop{\mathbb{E}}_{q_{\phi}(z_{<i}|x)} D_\KL\left[ q_\phi(z_i|x, z_{<i}) || p_\theta(z_i|z_{<i}) \right]
\end{equation}

\section{Related Work}
\label{related_work}
While HVAEs have been used on multiple occasions in literature \cite{ranganath2016hierarchical, sonderby2016ladder, klushyn2019learning}, they became especially prominent in recent works when these models became deep enough to compete with state-of-the-art autoregressive models. 

The most widely used base very deep HVAE models are NVAE \cite{vahdat2020nvae} and VDVAE \cite{child2021very}. During training, the former relies on heavy regularization in the hopes of stabilizing the unbounded KL terms of (\ref{helbo_eq}) and the latter -while achieving better NLL than NVAE- is less stable due to lack of regularization; both these models being more computationally expensive the deeper the model gets.

This work is based on VDVAE in the aim of reducing its computational requirements and adding more training stability. For completeness, we provide details of the architecture we use in all of our experiments in figure \ref{architecture_fig}. It is advisable to refer to the \href{https://github.com/Rayhane-mamah/Efficient-VDVAE}{source code} for more details.

\section{Revisiting VDVAE}
\label{contributions}

\subsection{Compute reduction} \label{compute_reduction}
\paragraph{Architecture design} VDVAE's work\cite{child2021very} empirically demonstrates that networks in general benefit from more layers at higher resolutions (measured in NLL), suggesting that it is important to have latent variables that learn local image details at high resolution layers. Our investigation shows that the benefit of adding layers to the high resolution layers is only noticeable when comparing NLL metrics and doesn't perceptively affect the model's reconstructed or generated images. The addition of high resolution layers also drastically increases the memory requirements of the networks. Moreover, the NLL gains from adding high resolutions layers follows the rule of diminishing returns. We show that it is possible to design memory efficient VDVAEs that don't affect NLL results.

\paragraph{Optimization} We study the effect of changing the optimization scheme in order to converge faster (fewer updates and faster clock time). We also train all our models with reduced batch sizes in order to save computational costs. These changes introduce more training instabilities. Therefore, we developed new methods to stabilize the models. 

\subsection{Stabilization} 
\paragraph{Gradient smoothing} One common problem with training deep Gaussian HVAEs is the very large gradients resulting from the unbounded KL term\cite{child2021very, vahdat2020nvae}, more specifically the gradients resulting from the inverse of the posterior standard deviation $1 / \sigma_{posterior}$. While NVAE attempts to solve this by using residual normal distributions and clipping the range of the $log(\sigma)$ predicted by the model, VDVAE counters this by clipping/skipping sharp gradients. In practice, we find that these solutions can hinder the model's learning depending on the dataset and depending on the model's other hyper-parameters. We propose instead to smooth the gradients of the inverse std by computing the Gaussian stds of both posteriors and priors with $\sigma = \mathrm{Softplus}(y, \beta) = \frac{1}{\beta} log(1 + exp(\beta y))$, with $y$ the activation of a linear layer and $\beta$ being a smoothing multiplier smaller than $1$. Like VDVAE, we use the discretized mixture of logistics (MoL) as the output layer of the generative model. To mitigate its sharp gradients, we similarly use gradient smoothing.

\paragraph{Optimizer} Since the norm of gradients of the MoL layer can be much larger than 1, especially when using small batch sizes, optimizing this model with the Adam optimizer \cite{kingma2017adam} can prove to be challenging since its second momentum $v_t = \beta_2 v_{t - 1} + (1 - \beta_2) g^2_t$ grows large. As such, we propose to use its infinite norm alternative, Adamax, which does not suffer from the same problem.

\begin{table}[]
\vskip 0.1in
\begin{subtable}{.49\textwidth}
\centering
\resizebox{.95\textwidth}{!}{%
\begin{tabular}{@{}ccccclcc@{}}
\toprule
\multicolumn{5}{c}{Distribution of Layers} &
   &
  \multirow{2}{*}{\begin{tabular}[c]{@{}c@{}}Memory \\ (GB)\end{tabular}} &
  \multirow{2}{*}{\begin{tabular}[c]{@{}c@{}}NLL\\ (bits/dim)\end{tabular}} \\ \cmidrule(){1-5}
$32 \times 32$ & $16 \times 16$ & $8 \times 8$ & $4 \times 4$ & $1 \times 1$ &  &       &       \\ \midrule
$11$ & $15$ & $19$ & $7$ & $6$ && $\bf{66}$ & $3.63$ \\
$16$ & $15$ & $19$ & $7$ & $6$ && $71$ & $3.61$ \\
$16$ & $25$ & $19$ & $7$ & $6$ && $76$ & $\bf{3.58}$ \\
$25$ & $25$ & $19$ & $7$ & $6$ && $88$ & $3.58$ \\
$41$ & $20$ & $10$ & $7$ & $6$ && $99$ & $3.59$ \\
\bottomrule
\end{tabular}
}
\subcaption{Distribution of layer depths.}
\label{depth_table}
\end{subtable}
\hfill
\begin{subtable}{.49\textwidth}
\centering
\resizebox{.95\textwidth}{!}{%
\begin{tabular}{@{}ccccclcc@{}}
\toprule
\multicolumn{5}{c}{Distribution of Layers} &
   &
  \multirow{2}{*}{\begin{tabular}[c]{@{}c@{}}Memory \\ (GB)\end{tabular}} &
  \multirow{2}{*}{\begin{tabular}[c]{@{}c@{}}NLL\\ (bits/dim)\end{tabular}} \\ \cmidrule(){1-5}
$32 \times 32$ & $16 \times 16$ & $8 \times 8$ & $4 \times 4$ & $1 \times 1$ &  &       &       \\ \midrule
$256$ & $256$ & $256$ & $256$ & $256$ && $64$ & $3.61$ \\
$32$ & $32$ & $64$ & $128$ & $256$ && $\bf{12}$ & $3.63$ \\
$64$ & $64$ & $128$ & $256$ & $512$ && $34$ & $3.61$ \\
$512$ & $512$ & $512$ & $512$ & $512$ && $99$ & $\bf{3.59}$ \\
\bottomrule
\end{tabular}
}
\subcaption{Distribution of layer widths.}
\label{width_table}
\end{subtable}
\caption{\textbf{NLL and memory load with different configurations of stochastic layers on ImageNet} $\bf{32\times32}$ (similar trends appear on the other benchmarks). \textbf{(a):} Networks with fixed low resolution, but with different number of middle and high resolution layers. The addition of extra layers in high resolution is a diminishing returns endeavor. \textbf{(b):} Networks with $84$ stochastic layers, but with varying layer widths. Reducing the width of the model can save memory while getting comparable NLL results.}
\end{table}

\begin{table}[]
\vskip 0.1in

\begin{subtable}{.57\textwidth}
\centering
\resizebox{.95\textwidth}{!}{
\begin{tabular}{@{}ccccccc@{}}
\toprule
\multirow{2}{*}{Batch size} &
  & \multicolumn{2}{c}{Without gradient smoothing} &
  & \multicolumn{2}{c}{With gradient smoothing} \\ \cmidrule(){3-4} \cmidrule(){6-7} & & NLL (bits/dim) & Skipped updates & & NLL (bits/dim) & Skipped updates \\ 
\midrule
$4$  & & $3.25$ & $78$ & & $3.22$ & $5$ \\ 
$8$  & & $3.11$ & $8$ & & $3.11$ & $4$ \\
$32$  & & $2.97$ & $3$ & & $2.97$ & $0$ \\
\bottomrule
\end{tabular}
}
\subcaption{Gradient smoothing and batch size.}
\end{subtable}
\hfill
\begin{subtable}{.42\textwidth}
\centering
\resizebox{\textwidth}{!}{
\begin{tabular}{@{}ccc@{}}
\toprule
\multirow{2}{*}{Dataset} & \multirow{2}{*}{\begin{tabular}[c]{@{}c@{}}Without \\ gradient smoothing \end{tabular}}
  & \multirow{2}{*}{\begin{tabular}[c]{@{}c@{}}With \\ gradient smoothing \end{tabular}} \\ \\
\midrule
FFHQ $1024\times1024$ & diverged & 6 \\ 
CelebAHQ $1024\times1024$ & diverged & 23 \\ 

\bottomrule
\end{tabular}
}
\subcaption{High resolution datasets.}
\label{extreme_gradient_smoothing_table}
\end{subtable}
\caption{\textbf{Effects of gradient smoothing}. (a) We trained networks with the same model architecture with varying the batch size for $500$k steps and tested the effect of gradient smoothing at $\beta=\log(2)$. The gradient smoothing does not affect NLL and results in greater stability when the batch size is reduced, as measured by the number of skipped updates at a  gradient update threshold of $800$ (for a negative ELBO measured in nats/dim). These results are computed on the CIFAR-10 dataset, but other datasets exhibit similar behaviors. (b) Number of skipped updates in $80$k steps on high resolution datasets for a threshold of 1200 (loss computed in nats/dim), $\beta$ = 0.4 and a batch size of $8$ . Gradient smoothing allows us to train models with small batch sizes.}
\label{batch_size_stability_table}
\end{table}

\section{Experiments}
\label{experiments}

\subsection{Model design, NLL and memory}\label{design_nll_mem}
We first tested to what extent increasing the depth of the model in the higher resolution layers is useful for the NLL metric, and how much memory can be saved by reducing this depth without hurting the model performance. In table \ref{depth_table}, we train multiple networks while fixing the number of low resolution layers and we experiment with only changing the number of medium and high resolution layers. We conclude that, after a certain depth, increasing the number of layers in high resolution latent spaces stops improving MLE performance. Distributing the number of layers across the lower resolution layers, not only saves memory, but also preserves performance. 

We then tested the impact of using the same big number of filters across all layers of the model against adopting an incremental strategy of the width, as can be seen in table \ref{width_table}. We show that high resolution latent spaces benefit less from width than the lower resolution latent spaces. We take advantage of this finding to considerably reduce the GPU memory load while remaining comparable to the baseline. In all of our experiments, we set a maximum GPU memory load of $320$GB.

\subsection{Gradient smoothing and stability}\label{grad_smooth_stabl}
Reducing the computation requirements by reducing the batch size as described in section \ref{compute_reduction} introduces training instabilities. We show in table \ref{batch_size_stability_table} that gradient smoothing substantially improves stability when using small batch sizes and that it is mandatory for high resolution datasets. For completeness, we show the effect of gradient smoothing as a function of model depth in Appendix \ref{depth_stability}.

\subsection{Quantitative model results}
\subsubsection{Computation load comparison}
We start by examining the training memory load gain between our Efficient-VDVAE and the original VDVAE. All VDVAE parameters are reported in the original VDVAE work\footnote{On occasions, the official released codebase parameters differs from what's reported in the paper. In that case we rely on the codebase version.}, and we train our models with only applying the architecture and optimization modifications described in section \ref{contributions} and tested in experiments \ref{design_nll_mem}, \ref{grad_smooth_stabl}.

\begin{table}[]
\centering
\caption{\textbf{Computational load comparison}. We train $2$ configurations of Efficient-VDVAE, derived from the VDVAE baseline by only modifying the hyper-parameters described in section \ref{contributions}. We report the convergence speed as measured in number of iterations and clock time (in hours). We also report the total training memory load of each model and their NLL at convergence time. The configuration C1 is only different from C2 in the use of incremental filter width (experiment of table \ref{width_table}). A south-east arrow ($\searrow$) denotes the use of cosine decay. More detailed model hyper-parameters are available in table \ref{layer_dist} and the \href{https://github.com/Rayhane-mamah/Efficient-VDVAE}{source code}.}
\vskip 0.1in
\resizebox{\textwidth}{!}{%
\begin{tabular}{@{}cccclccclccclccc@{}}
\toprule
Dataset &
\multicolumn{3}{c}{CIFAR-10} &
&
\multicolumn{3}{c}{\begin{tabular}[c]{@{}c@{}}Imagenet $32\times32$\end{tabular}} &
&
\multicolumn{3}{c}{\begin{tabular}[c]{@{}c@{}}Imagenet $64\times64$\end{tabular}} &
&
\multicolumn{3}{c}{\begin{tabular}[c]{@{}c@{}}FFHQ $256\times256$ (5-bits)\end{tabular}} \\ \cmidrule(){2-4} \cmidrule(){6-8} \cmidrule(){10-12} \cmidrule(){14-16} 
Model & C1 & C2 & VDVAE &  & C1 & C2 & VDVAE &  & C1 & C2 & VDVAE &  & C1 & C2 & VDVAE \\ 
\midrule
Layers & $47$ & $47$ & $43$ &  & $73$ & $73$ & $78$ &  & $84$ & $84$ & $75$ &  & $66$ & $66$ & $66$ \\
width & incr.  & $384$ & $384$ &  & incr.  & $512$ & $512$ &  & incr.  & $512$ & $512$ &  & incr.  & $512$ & $512$ \\
Batch size & $16$ & $16$ & $32$ &  & $64$ & $64$ & $256$ &  & $32$ & $32$ & $128$ &  & $16$ & $16$ & $32$ \\
Learning rate & $0.001\searrow$ & $0.001\searrow$ & $0.0001$ &  & $0.001\searrow$ & $0.001\searrow$ & $0.0001$ &  & $0.001\searrow$ & $0.001\searrow$ & $0.0001$ &  & $0.001\searrow$ & $0.001\searrow$ & $0.0001$ \\
Optimizer & Adamax & Adamax & Adam &  & Adamax & Adamax & Adam & & Adamax & Adamax  & Adam &  & Adamax & Adamax & Adam  \\ 
Parameters & $18$M & $57$M & $39$M &  & $52$M & $145$M & $119$M & & $57$M & $168$M  & $125$M &  & $67$M & $198$M & $115$M  \\ 
\midrule
Time (h) & $\bf{99}$ & $108$ & $144$ &  & $\bf{161}$ & $177$ & $420$ &  & $\bf{184}$ & $202$ & $420$ &  & $\bf{226}$ & $245$ & $420$ \\
Training iter. & $800$k & $800$k & $1.1$M &  & $800$k & $800$k & $1.7$M &  & $800$k & $800$k & $1.6$M &  & $850$k & $850$k & $1.7$M  \\
Memory (GB) & $\bf{8}$ & $17$ & $32$ &  & $\bf{26}$ & $76$ & $512$ &  & $\bf{29}$ & $116$ & $512$ &  & $\bf{101}$ & $304$ & $512$ \\
NLL (bits/dim) & $\leq 2.91$ & $\leq \bf{2.87}$ & $\leq \bf{2.87}$ &  & $\leq 3.60$ & $\leq \bf{3.58}$ & $\leq 3.80$ &  & $\leq 3.33$ & $\leq \bf{3.30}$ & $\leq 3.52$ &  & $\leq 0.61$ & $\leq \bf{0.60}$ & $\leq 0.61$ \\
\bottomrule
\end{tabular}%
\label{quantitative_compute}
}
\end{table}

In table \ref{quantitative_compute}, we report the results on CIFAR-10 \cite{cifar}, ImageNet $32\times32$, ImageNet $64\times64$ \cite{deng2009imagenet} and FFHQ $256\times256$ \cite{ffhq} datasets. In the worst case, Efficient-VDVAE has a $1.68\times$ lower training memory load, converges in $1.38\times$ less updates and trains $1.45\times$ faster in clock time, while being consistently similar or better than the original VDVAE in terms of NLL. More compact Efficient-VDVAE configurations can also be trained to consume $20\times$ less memory load, to converge in $2\times$ less updates and to train in $2.6\times$ less clock time on Imagenet $32\times 32$. In this scenario, computational gains can worsen the NLL metric by up to $0.02\%$ which is usually qualitatively indistinguishable.

\subsubsection{NLL results}
In table \ref{nll_sota_table}, we report the NLL scores compared to other notable MLE models. We report these results on more datasets than in the original VDVAE for completeness: We add NLL results on MNIST\cite{lecun-mnisthandwrittendigit-2010}, CelebA $64\times64$\cite{liu2015deep, pmlr-v48-larsen16}, CelebAHQ $256\times256$ and CelebAHQ $1024\times1024$\cite{celebahq}. Most of these additions are commonly used except CelebAHQ $1024\times1024$ which we added to act as a baseline for high resolution images for future work. We also provide the 8-bit scores on FFHQ $256\times256$ and CelebAHQ $256\times256$ which are usually trained in 5-bits (see section \ref{lossless_compression} for an in-depth explanation).

Despite being more efficient than VDVAE, Efficient-VDVAE consistently achieves either similar or better NLL scores on all baselines. It also achieves state-of-the-art performance on all datasets but CIFAR-10, although it remains comparable to non-augmented models. See \href{https://github.com/Rayhane-mamah/Efficient-VDVAE}{codebase} for a rundown of all hyper-parameters used for each dataset. 

\begin{table}[]
\centering
\caption{\textbf{Comparison against state-of-the-art likelihood-based generative models.} Unless otherwise specified, all datasets have 8-bit images. All benchmarks are in bits/dim except for MNIST benchmarks which are reported in nats. An asterisk (*) denotes scores with data augmentation. Since most models overfit on CIFAR-10, data augmentation greatly improves performance. Our computational efficiency allows us to set HVAE baselines on high resolution datasets. Efficient-VDVAE outperforms current state-of-the-art models on most datasets.}
\vskip 0.1in
\label{nll_sota_table}
\resizebox{\textwidth}{!}{%
\begin{tabular}{@{}ccccccccccccccc@{}}
\toprule
\multirow{2}{*}{Model} &
  \multirow{2}{*}{MNIST} &
  \multirow{2}{*}{CIFAR-10} &
  \multirow{2}{*}{\begin{tabular}[c]{@{}c@{}}Imagenet\\ $32\times32$\end{tabular}} &
  \multirow{2}{*}{\begin{tabular}[c]{@{}c@{}}Imagenet\\ $64\times64$\end{tabular}} &
  \multirow{2}{*}{\begin{tabular}[c]{@{}c@{}}CelebA\\ $64\times64$\end{tabular}} &
  \multirow{2}{*}{} &
  \multicolumn{2}{c}{\begin{tabular}[c]{@{}c@{}}CelebAHQ\\ $256\times256$\end{tabular}} &
  \multirow{2}{*}{} &
  \multicolumn{2}{c}{\begin{tabular}[c]{@{}c@{}}FFHQ\\ $256\times256$\end{tabular}} &
  \multirow{2}{*}{} &
  \multirow{2}{*}{\begin{tabular}[c]{@{}c@{}}CelebAHQ\\ $1024\times1024$\end{tabular}} &
  \multirow{2}{*}{\begin{tabular}[c]{@{}c@{}}FFHQ\\ $1024\times1024$\end{tabular}} \\ \cmidrule(){8-9} \cmidrule(){11-12}
 &  &  &  &  &  &  & 5-bits & 8-bits &  & 5-bits & 8-bits &  &  &  \\ 
 \midrule
ANF\cite{anf} & - & $3.05$ & $3.92$ & - & - &  & $0.72$ & - &  & - & - &  & - & - \\
Flow++\cite{flow++} & - & $3.08$ & $3.86$ & $3.69$ & - &  & - & - &  & - & - &  & - & - \\
GLOW\cite{kingma2018glow} & - & $3.35$ & $4.09$ & $3.81$ & - &  & $1.03$ & - &  & - & - &  & - & - \\
DenseFlow\cite{denseflow} & - & $2.98$ & $3.63$ & $3.35$ & $1.99$ &  & - & - &  & - & - &  & - & - \\
\midrule
$\delta$-VAE\cite{delta_vae} & - & $2.83$ & $3.77$ & - & - &  & - & - &  & - & - &  & - & - \\
SPN\cite{spn} & - & - & $3.85$ & $3.52$ & - &  & $0.61$ & - &  & - & - &  & - & - \\
MaCow\cite{ma2019macow} & - & $3.16$ & - & $3.69$ & - &  & $0.67$ & - &  & - & - &  & - & - \\
PixelVAE++\cite{sadeghi2019pixelvae++} & $78.00$ & $2.90$ & - & - & - &  & - & - &  & - & - &  & - & - \\
Locally Masked PixelCNN\cite{locally} & $\bf{77.58}$ & $2.89$ & - & - & - &  & $0.74$ & - &  & - & - &  & - & - \\
Image Transformer\cite{image} & - & $2.89$ & $3.77$ & - & - &  & - & - &  & - & - &  & - & - \\
Sparse Transformer\cite{child2019generating} & - & $\bf{2.80}$ & - & $3.44$ & - &  & - & - &  & - & - &  & - & - \\
Aug. Sparse Transformer\cite{augmented_sparse_transformer} & - & $2.53$* & - & - & - &  & - & - &  & - & - &  & - & - \\
\midrule
UDM\cite{udm_paper} & - & $3.04$ & - & - & $1.93$ &  & - & - &  & - & - &  & - & - \\
VDM\cite{kingma2021variational} & - & $\leq \bf{2.49}$* & $\leq 3.72$* & $\leq 3.40$* & - &  & - & - &  & - & - &  & - & - \\
\midrule
NVAE\cite{vahdat2020nvae} & $\leq 78.01$ & $\leq 2.91$ & $\leq 3.92$ & - & $\leq 2.03$ &  & $\leq 0.70$ & - &  & $\leq 0.69$ & - &  & - & - \\
VDVAE\cite{child2021very} & - & $\leq 2.87$ & $\leq 3.80$ & $\leq 3.52$ & - &  & - & - &  & $\leq 0.61$ & - &  & - & $\leq 2.42$ \\
CR-NVAE\cite{sinha2021consistency} & $\leq \bf{76.93}$* & $\leq 2.51$* & - & - & $\leq 1.86$* &  & - & - &  & - & - &  & - & - \\
\midrule
Efficient-VDVAE & - & $\leq 2.87$ & $\leq \bf{3.58}$ & $\leq \bf{3.30}$ & $\leq \bf{1.83}$ &  & $\leq \bf{0.57}$ & - &  & $\leq \bf{0.60}$ & - &  & - & - \\
Efficient-VDVAE (section \ref{lossless_compression}) & $\leq 79.09$ & $\leq 2.87$ & $\leq \bf{3.58}$ & $\leq \bf{3.30}$ & $\leq \bf{1.83}$ &  & $\leq \bf{0.51}$ & $\leq \bf{1.35}$ &  & $\leq \bf{0.53}$ & $\leq \bf{2.17}$ &  & $\leq \bf{1.01}$ & $\leq \bf{2.30}$ \\ 
\bottomrule
\end{tabular}%
}
\end{table}

\begin{figure}
    \centering
    \begin{subfigure}[t]{0.32\textwidth}
        \centering
        \includegraphics[width=\linewidth]{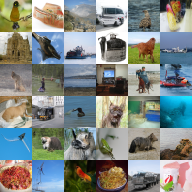} 
        \caption{Imagenet $32\times32$ ($t = 0.85$).}
    \end{subfigure}
    \hfill
    \begin{subfigure}[t]{0.32\textwidth}
        \centering
        \includegraphics[width=\linewidth]{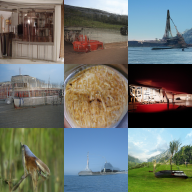} 
        \caption{Imagenet $64\times64$ ($t = 0.9$).}
    \end{subfigure}
    \hfill
    \begin{subfigure}[t]{0.32\textwidth}
        \centering
        \includegraphics[width=\linewidth]{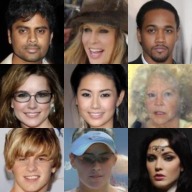} 
        \caption{CelebA $64\times64$ ($t = 0.8$).}
    \end{subfigure}
    \vfill
    \begin{subfigure}[t]{.425\textwidth}
    \centering
        \includegraphics[width=\linewidth]{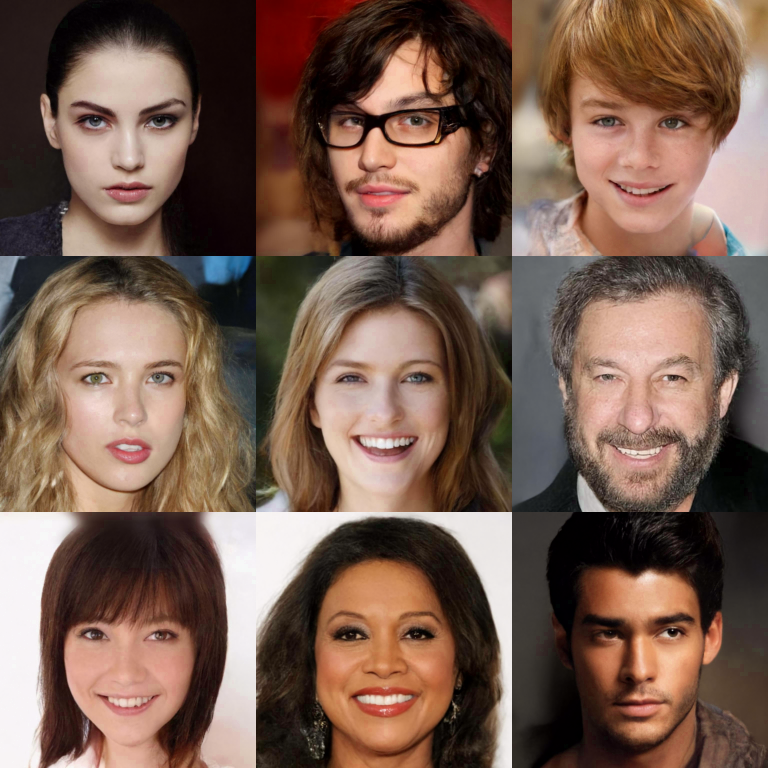} 
        \caption{CelebAHQ $256\times256$ 5-bits ($t = 0.85$).}
    \end{subfigure}
    \hskip 0.1in
    \begin{subfigure}[t]{.425\textwidth}
    \centering
        \includegraphics[width=\linewidth]{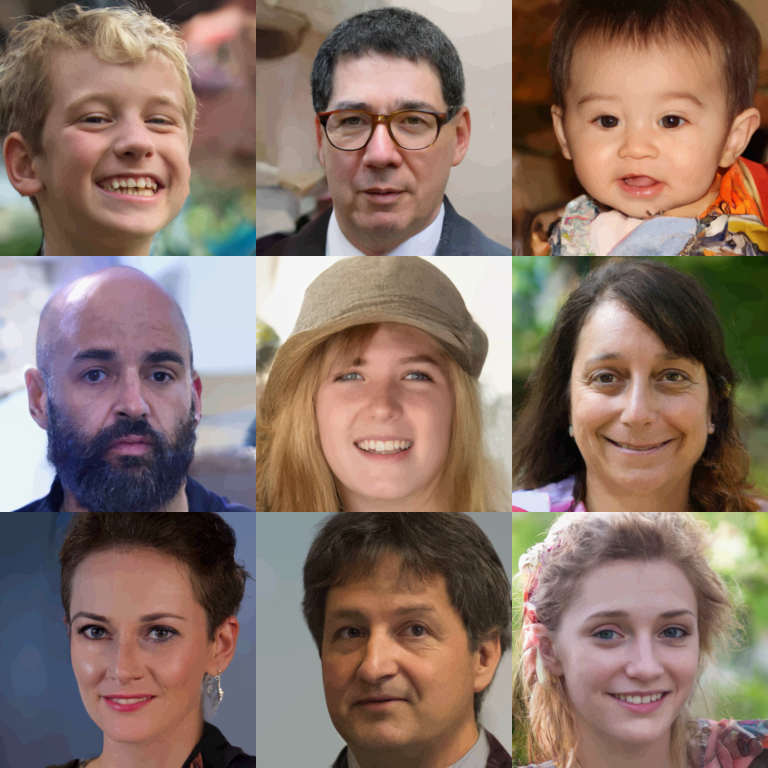} 
        \caption{FFHQ $256\times256$ 5-bits ($t = 0.8$).}
    \end{subfigure}
    \vfill
    \begin{subfigure}[t]{.425\textwidth}
    \centering
        \includegraphics[width=\linewidth]{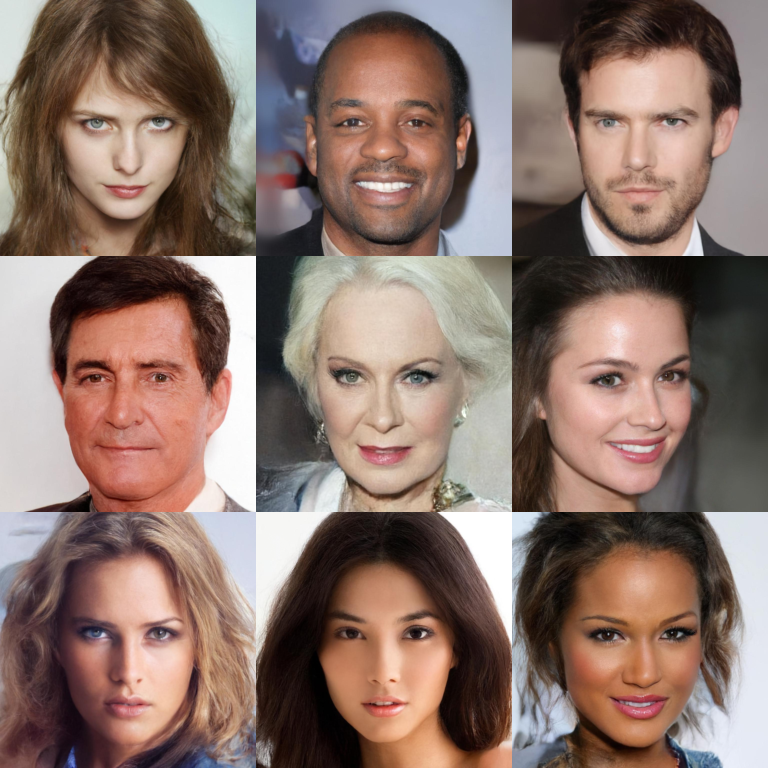} 
        \caption{CelebAHQ $256\times256$ 8-bits ($t = 0.85$).}
    \end{subfigure}
    \hskip 0.1in
    \begin{subfigure}[t]{.425\textwidth}
    \centering
        \includegraphics[width=\linewidth]{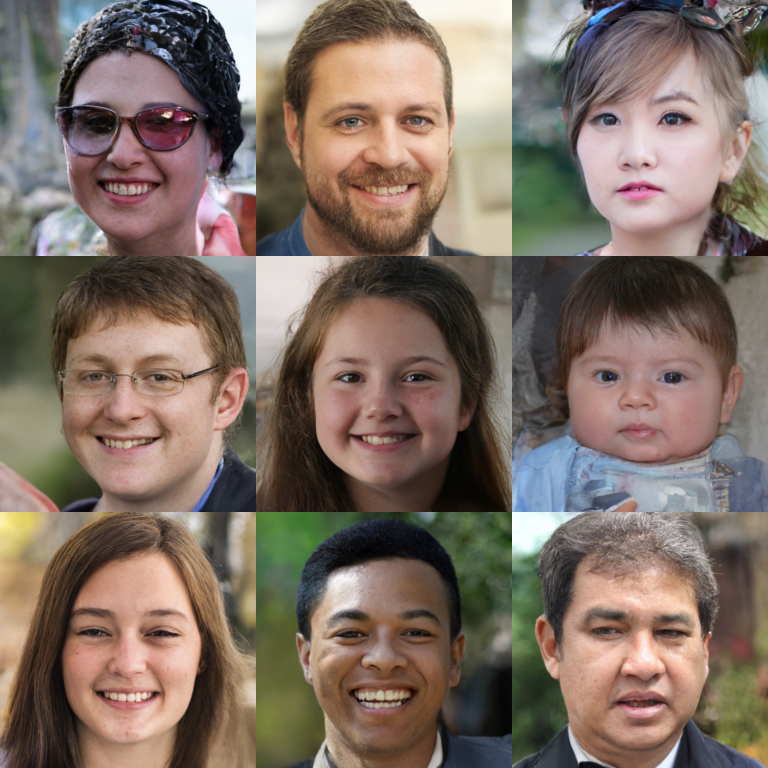} 
        \caption{FFHQ $256\times256$ 8-bits ($t = 0.8$).}
    \end{subfigure}
    \caption{\textbf{Samples generated from the prior of Efficient-VDVAE with temperature ($\bf{t}$)} (best viewed zoomed in). Compared to 5-bit images, the full depth color images don't have banding effects. They also introduce small details missing in 5-bit images (moles, freckles, light reflections, etc.). More samples are available in appendix \ref{additional_prior_samples_appendix}.}
    \label{prior_samples}
\end{figure}

\subsection{Qualitative model results}
Figure \ref{prior_samples} illustrates generated samples from the prior using the equation \ref{generative_model_eq}. We show a comparable subjective image quality and diversity to that of VDVAE. Although our models achieve a substantial improvement on the FFHQ $1024\times1024$ NLL compared to the VDVAE baseline, they are still not expressive enough to generate good high quality high resolution images. We show unconditional samples with different temperatures on FFHQ $1024\times1024$ and CelebAHQ $1024\times1024$ in \ref{additional_prior_samples_appendix}.

\section{Information theoretic interpretation and output layer dilemma}
\label{lossless_compression}
\subsection{Lossless compression}

From an information theory perspective, the reconstruction loss is interpreted as the term that copies information from the pixel space to the latent space, and the KL divergence is interpreted as the regularization term that compresses that information\cite{yu2020frombayestoit, chen2016variational}. Thus, during training, we expect the reconstruction loss to decrease and the KL divergence to increase as more information gets stored in the latent space. When the reconstruction loss reaches a minimum at which it can no longer decrease, we expect the KL divergence term to decrease as information in the latent space gets compressed.

While empirically desirable, VAEs (and HVAEs) rarely achieve perfect reconstructions. This is in part due to a pre-existing limit on the sharpness of the output layer; the MoL layer in the case of VDVAE. While this has been originally implemented to preserve training stability\cite{salimans2017pixelcnn++}, in our experiments it proved to be unnecessary after the gradient smoothing addition. We study the effect of removing such a bound in the next section.

\begin{figure}
    \centering
    \includegraphics[width=\textwidth]{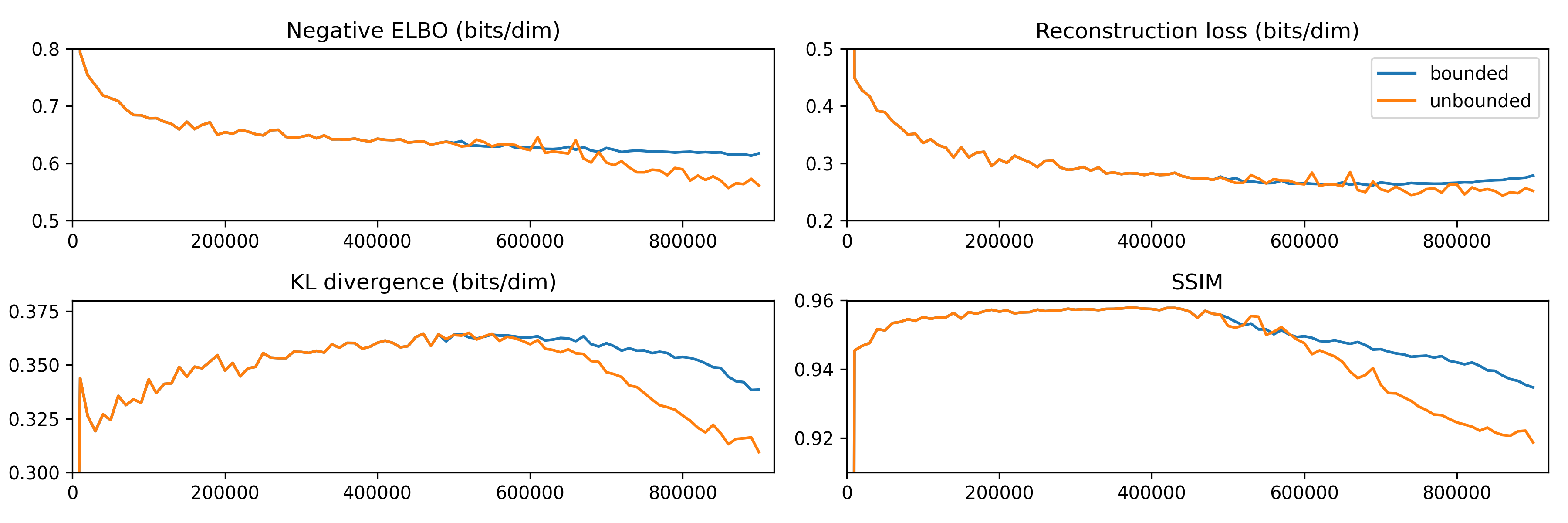}
    \caption{\textbf{Impact of the bound on metrics}: We compare the metrics of the bounded model and the unbounded model both trained on FFHQ $256\times256$. We observe that the unbounded model has a lower reconstruction loss. We also observe that the KL divergence converges to a lower value which then allows the Negative ELBO to reach a better final value. This however does not reflect the overall closeness of the reconstructed images to the target images as demonstrated by the Structural Similarity Index Measure (SSIM)\cite{Wang2004ImageQA} values.}
    \label{costs_fig}
\end{figure}

\subsection{Impact on 5-bit quantization}

The removal of the bound on the MoL layer allows us to adhere more accurately to the principles of lossless compression, and more importantly, avoid training our networks in an over-regularized regime\cite{yu2020frombayestoit}. In table \ref{nll_sota_table}, we test the effect of this removal on FFHQ $256\times256$ and CelebAHQ $256\times256$ which have been consistently quantized to 5-bits in prior work \cite{child2021very, vahdat2020nvae, 5bits_1, 5bits_2, ma2019macow, spn, anf, locally, kingma2018glow}.
We achieve much better NLL than state-of-the-art models on these 5-bit benchmarks. We report in figure \ref{costs_fig} the evolution of the validation metrics of a bounded and an unbounded model on FFHQ $256\times256$ during training.

\begin{figure}[t]
\begin{floatrow}

\ffigbox{
\includegraphics[width=.4\textwidth]{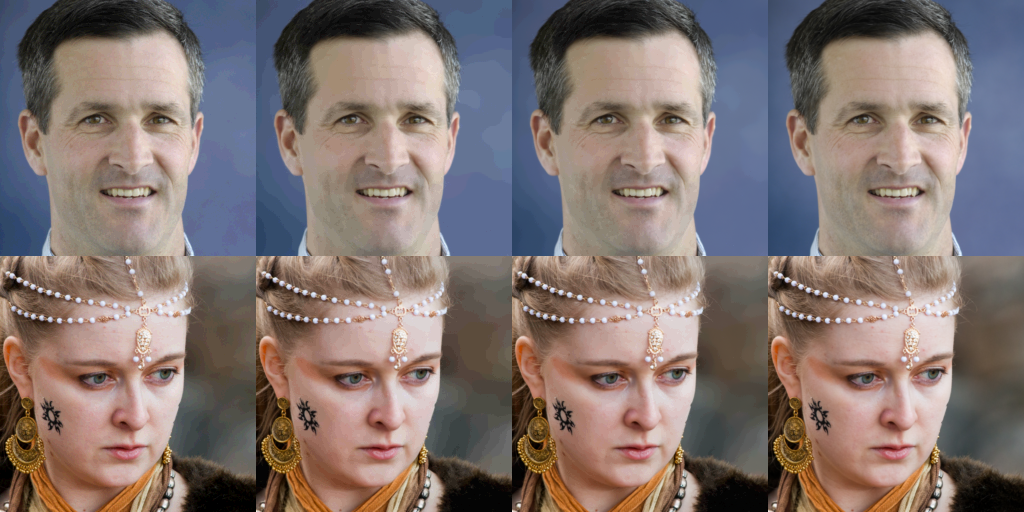}
}{
\caption{5-bit quantization and banding effects on FFHQ $256\times256$ (best viewed zoomed in). From left to right: 5-bit targets, unbounded reconstruction, bounded reconstruction and finally the 8-bit targets. All on FFHQ 256 ×256.}
\label{5bits_8bits_comparison}
}
\capbtabbox{
\resizebox{.46\textwidth}{!}{%
\begin{tabular}{@{}ccccc@{}}
\toprule
\begin{tabular}[c]{@{}c@{}}$\%$ of used \\ posteriors\end{tabular} &
\begin{tabular}[c]{@{}c@{}}Encoded dataset \\ size (GB) \end{tabular} &
\begin{tabular}[c]{@{}c@{}}Negative\\ ELBO\end{tabular} & Reconstruction & KL divergence \\ 
\midrule

$2.5\%$ & $111$ & $2.43$ & $1.16$  & $1.27$ \\
$\bf{3\boldsymbol{\%}}$ & $\bf{133}$ & $\bf{2.20}$ & $\bf{1.05}$ & $\bf{1.15}$ \\
$4\%$ & $177$ & $2.18$ & $1.05$ & $1.13$ \\ 
$5\%$ & $222$ & $2.17$ & $1.04$ & $1.13$  \\  
$7\%$ & $310$ & $2.17$ & $1.04$ & $1.13$  \\ 
$100\%$ & $4433$ & $2.17$ & $1.04$ & $1.13$  \\ 
\bottomrule
\end{tabular}%
}
}{
\caption{Evolution of the encoded dataset ($z$) size and test costs (bits/dim) with the pruning of the latent posteriors on 8-bit FFHQ $256\times256$ (uncompressed pixels size is $\approx 50$GB). We observe that by only using $3\%$ of the latent space, the encoded dataset size can be $33\times$ smaller with a comparable negative ELBO results.}
\label{nll_vs_percent}
}
\end{floatrow}
\end{figure}

Qualitatively, the model appears to compress the latent space information attributed to the banding effect \cite{banding_1, banding_2} which is heavily present in the 5-bit target images. Figure \ref{5bits_8bits_comparison} illustrates the appearance of banding effects in the 5-bit targets, and their effect on the unbounded models' reconstructions. We show differences between unconditional generated samples in 8-bits and 5-bits in figure \ref{prior_samples}.

Although prior work\cite{kingma2018glow} has introduced the use of the 5-bit benchmarks as a means to reduce the datasets' complexity by removing the high frequency color depth information, current state-of-the-art models are more capable of modeling the high frequency noise. Since VAEs are trained to mimic the training data distribution, and, any bias introduced in the data preparation makes them generate samples with a similar bias, they are expected to mimic the banding effect which is undesirable from a generative modeling perspective. Additionally, as HVAEs get more expressive, NLL differences on 5-bit datasets start to only contrast how well the models are compressing the banding effects, rather than giving a global sense of how well the HVAEs are learning the data distribution. Thus, we advocate that all future work should be evaluated on the 8-bit images. As a step in this direction, we provide in table \ref{nll_sota_table} NLL scores for the 8-bit version of the FFHQ $256\times256$ and CelebAHQ $256\times256$ datasets.

Lastly, we observe in table \ref{nll_sota_table}, that the removal of the output bound does not affect the NLL results of the 8-bit benchmarks. We hypothesize that this is because our current models are not expressive enough to produce the same KL compression effect on full color depth images.

\section{Effective latent space}
\label{effective_latent_space}

In their dense form, $z$ can occupy a lot of memory, especially with very deep HVAEs, which is not suitable for downstream tasks. Fortunately, in practice, VAEs (and HVAEs) train in a polarized regime and learn a latent space where the majority of the latent dimensions are "turned off" for all samples \cite{rolinek2019variational, Duan2020Unsupervised}.

We experiment with "pruning" the posteriors that are "turned-off" by replacing them with the prior and observe its effect on the test scores. By definition of the polarized regime\cite{rolinek2019variational}, dimensions are turned-off when their average KL divergence (independent from the latent resolution) is lower than a certain threshold. We report, in table \ref{nll_vs_percent}, the effect of moving the cutoff threshold on CelebAHQ $256\times256$ (these results hold for other datasets as well). 

We discover that roughly $3\%$ of the latent dimensions encode most of the information required to reconstruct the inputs as shown by an almost equal reconstruction loss. The remaining $97\%$ posterior dimensions can be collapsed on the prior without any qualitative effects as seen in figure \ref{3percent_reconstructions_fig}. Therefore, we theorize that using the "active dimensions" of the latent space alone could be sufficient for downstream tasks, while also being memory efficient.

\section{Limitations and broader impact}
\label{broader_impact}

This paper's contributions are based on the fundamental need for a cheaper and stabler version of the Deep HVAE. The contributions of this paper have been tested on commonly used datasets for a full comparison with current state-of-the-art.

This work makes it more accessible to build applications in representation learning, content generation and semi-supervised learning. However, these applications are known to have potential negative societal impacts, as the model's representation could be biased and can harm the livelihood of minorities and marginalized groups. Making these models more accessible can also simplify harmful content generation. More work needs to be done on these models to eliminate such deficiencies.

Despite our work, HVAEs still require a relatively large amount of compute, especially on high resolution datasets, which can still make them inaccessible for individuals and small groups. From a stability point of view, very deep HVAEs that use unimodal latent distributions with infinite support remain, to some extent, unstable by design. Using different latent distributions \cite{impact1, impact2, impact3, impact4} may prove fruitful for completely stabilizing HVAEs. However, using different latent distributions while achieving the same NLL performance of the Gaussian distribution is still an open area of research.

\section{Conclusion}
\label{conclusion}

In this paper, we show that VDVAE's stability and computational efficiency could be improved while retaining or boosting the MLE performance as measured in NLL. We then argue against benchmarking on 5-bit quantized datasets, and we reason that future work should be evaluated on full color depth datasets. We also empirically show that only a minor percentage of HVAE's sparse latent dimensions is responsible for encoding most of the information. In the spirit of making these models accessible, we have publicly released our \href{https://github.com/Rayhane-mamah/Efficient-VDVAE}{source code}.

\section*{Acknowledgments}
The authors would like to thank Alex Krizhevsky for his mentorship and insightful discussions. The authors also thank Joe Palermo, Marc Tyndel, Hashiam Kadhim and  Stephen Piron for their feedback and support. 

\bibliographystyle{unsrt}  
\bibliography{references}

\begin{thebibliography}{10}

\bibitem{van2016conditional}
Aaron Van~den Oord, Nal Kalchbrenner, Lasse Espeholt, Oriol Vinyals, Alex
  Graves, et~al.
\newblock Conditional image generation with pixelcnn decoders.
\newblock {\em Advances in neural information processing systems}, 29, 2016.

\bibitem{salimans2017pixelcnn++}
Tim Salimans, Andrej Karpathy, Xi~Chen, and Diederik~P Kingma.
\newblock Pixelcnn++: Improving the pixelcnn with discretized logistic mixture
  likelihood and other modifications.
\newblock {\em arXiv preprint arXiv:1701.05517}, 2017.

\bibitem{sadeghi2019pixelvae++}
Hossein Sadeghi, Evgeny Andriyash, Walter Vinci, Lorenzo Buffoni, and
  Mohammad~H Amin.
\newblock Pixelvae++: Improved pixelvae with discrete prior.
\newblock {\em arXiv preprint arXiv:1908.09948}, 2019.

\bibitem{augmented_sparse_transformer}
Heewoo Jun, Rewon Child, Mark Chen, John Schulman, Aditya Ramesh, Alec Radford,
  and Ilya Sutskever.
\newblock Distribution augmentation for generative modeling.
\newblock In Hal~Daumé III and Aarti Singh, editors, {\em Proceedings of the
  37th International Conference on Machine Learning}, volume 119 of {\em
  Proceedings of Machine Learning Research}, pages 5006--5019. PMLR, 13--18 Jul
  2020.

\bibitem{kingma2014autoencoding}
Diederik~P Kingma and Max Welling.
\newblock Auto-encoding variational bayes, 2014.

\bibitem{rezende2014stochastic}
Danilo~Jimenez Rezende, Shakir Mohamed, and Daan Wierstra.
\newblock Stochastic backpropagation and approximate inference in deep
  generative models, 2014.

\bibitem{bengio_rl}
Yoshua Bengio, Aaron~C. Courville, and Pascal Vincent.
\newblock Unsupervised feature learning and deep learning: {A} review and new
  perspectives.
\newblock {\em CoRR}, abs/1206.5538, 2012.

\bibitem{radford_rl}
Alec Radford, Luke Metz, and Soumith Chintala.
\newblock Unsupervised representation learning with deep convolutional
  generative adversarial networks.
\newblock {\em arXiv preprint arXiv:1511.06434}, 2015.

\bibitem{vae_diversity}
Sam Bond{-}Taylor, Adam Leach, Yang Long, and Chris~G. Willcocks.
\newblock Deep generative modelling: {A} comparative review of vaes, gans,
  normalizing flows, energy-based and autoregressive models.
\newblock {\em CoRR}, abs/2103.04922, 2021.

\bibitem{diffusion_1}
Jonathan Ho, Ajay Jain, and Pieter Abbeel.
\newblock Denoising diffusion probabilistic models.
\newblock {\em Advances in Neural Information Processing Systems},
  33:6840--6851, 2020.

\bibitem{diffusion_2}
Jascha Sohl-Dickstein, Eric Weiss, Niru Maheswaranathan, and Surya Ganguli.
\newblock Deep unsupervised learning using nonequilibrium thermodynamics.
\newblock In {\em International Conference on Machine Learning}, pages
  2256--2265. PMLR, 2015.

\bibitem{diffusion_3}
Alexander~Quinn Nichol and Prafulla Dhariwal.
\newblock Improved denoising diffusion probabilistic models.
\newblock In {\em International Conference on Machine Learning}, pages
  8162--8171. PMLR, 2021.

\bibitem{gan1}
Ian Goodfellow, Jean Pouget-Abadie, Mehdi Mirza, Bing Xu, David Warde-Farley,
  Sherjil Ozair, Aaron Courville, and Yoshua Bengio.
\newblock Generative adversarial nets.
\newblock {\em Advances in neural information processing systems}, 27, 2014.

\bibitem{gan2}
Phillip Isola, Jun{-}Yan Zhu, Tinghui Zhou, and Alexei~A. Efros.
\newblock Image-to-image translation with conditional adversarial networks.
\newblock {\em CoRR}, abs/1611.07004, 2016.

\bibitem{gan3}
Jun{-}Yan Zhu, Taesung Park, Phillip Isola, and Alexei~A. Efros.
\newblock Unpaired image-to-image translation using cycle-consistent
  adversarial networks.
\newblock {\em CoRR}, abs/1703.10593, 2017.

\bibitem{child2021very}
Rewon Child.
\newblock Very deep {\{}vae{\}}s generalize autoregressive models and can
  outperform them on images.
\newblock In {\em International Conference on Learning Representations}, 2021.

\bibitem{kingma2019vae_tutorial}
Diederik~P. Kingma and Max Welling.
\newblock An introduction to variational autoencoders.
\newblock {\em CoRR}, abs/1906.02691, 2019.

\bibitem{yu2020frombayestoit}
Ronald Yu.
\newblock A tutorial on vaes: From bayes' rule to lossless compression.
\newblock {\em CoRR}, abs/2006.10273, 2020.

\bibitem{ranganath2016hierarchical}
Rajesh Ranganath, Dustin Tran, and David Blei.
\newblock Hierarchical variational models.
\newblock In {\em International conference on machine learning}, pages
  324--333. PMLR, 2016.

\bibitem{NIPS2016_ddeebdee}
Durk~P Kingma, Tim Salimans, Rafal Jozefowicz, Xi~Chen, Ilya Sutskever, and Max
  Welling.
\newblock Improved variational inference with inverse autoregressive flow.
\newblock In D.~Lee, M.~Sugiyama, U.~Luxburg, I.~Guyon, and R.~Garnett,
  editors, {\em Advances in Neural Information Processing Systems}, volume~29.
  Curran Associates, Inc., 2016.

\bibitem{sonderby2016ladder}
Casper~Kaae S{\o}nderby, Tapani Raiko, Lars Maal{\o}e, S{\o}ren~Kaae
  S{\o}nderby, and Ole Winther.
\newblock Ladder variational autoencoders.
\newblock {\em Advances in neural information processing systems}, 29, 2016.

\bibitem{klushyn2019learning}
Alexej Klushyn, Nutan Chen, Richard Kurle, Botond Cseke, and Patrick van~der
  Smagt.
\newblock Learning hierarchical priors in vaes.
\newblock {\em Advances in neural information processing systems}, 32, 2019.

\bibitem{vahdat2020nvae}
Arash Vahdat and Jan Kautz.
\newblock Nvae: A deep hierarchical variational autoencoder.
\newblock {\em Advances in Neural Information Processing Systems},
  33:19667--19679, 2020.

\bibitem{kingma2017adam}
Diederik~P. Kingma and Jimmy Ba.
\newblock Adam: A method for stochastic optimization, 2017.

\bibitem{cifar}
Alex Krizhevsky, Vinod Nair, and Geoffrey Hinton.
\newblock Cifar-10 (canadian institute for advanced research).

\bibitem{deng2009imagenet}
Jia Deng, Wei Dong, Richard Socher, Li-Jia Li, Kai Li, and Li~Fei-Fei.
\newblock Imagenet: A large-scale hierarchical image database.
\newblock In {\em 2009 IEEE conference on computer vision and pattern
  recognition}, pages 248--255. Ieee, 2009.

\bibitem{ffhq}
Tero Karras, Samuli Laine, and Timo Aila.
\newblock Flickr faces hq (ffhq) 70k from stylegan.
\newblock {\em CoRR}, 2018.

\bibitem{lecun-mnisthandwrittendigit-2010}
Yann LeCun and Corinna Cortes.
\newblock {MNIST} handwritten digit database.
\newblock 2010.

\bibitem{liu2015deep}
Ziwei Liu, Ping Luo, Xiaogang Wang, and Xiaoou Tang.
\newblock Deep learning face attributes in the wild.
\newblock In {\em Proceedings of the IEEE international conference on computer
  vision}, pages 3730--3738, 2015.

\bibitem{pmlr-v48-larsen16}
Anders Boesen~Lindbo Larsen, Søren~Kaae Sønderby, Hugo Larochelle, and Ole
  Winther.
\newblock Autoencoding beyond pixels using a learned similarity metric.
\newblock In Maria~Florina Balcan and Kilian~Q. Weinberger, editors, {\em
  Proceedings of The 33rd International Conference on Machine Learning},
  volume~48 of {\em Proceedings of Machine Learning Research}, pages
  1558--1566, New York, New York, USA, 20--22 Jun 2016. PMLR.

\bibitem{celebahq}
Tero Karras, Timo Aila, Samuli Laine, and Jaakko Lehtinen.
\newblock Progressive growing of gans for improved quality, stability, and
  variation.
\newblock {\em CoRR}, abs/1710.10196, 2017.

\bibitem{anf}
Chin-Wei Huang, Laurent Dinh, and Aaron Courville.
\newblock Augmented normalizing flows: Bridging the gap between generative
  flows and latent variable models.
\newblock {\em arXiv preprint arXiv:2002.07101}, 2020.

\bibitem{flow++}
Jonathan Ho, Xi~Chen, Aravind Srinivas, Yan Duan, and Pieter Abbeel.
\newblock Flow++: Improving flow-based generative models with variational
  dequantization and architecture design.
\newblock In {\em International Conference on Machine Learning}, pages
  2722--2730. PMLR, 2019.

\bibitem{kingma2018glow}
Durk~P Kingma and Prafulla Dhariwal.
\newblock Glow: Generative flow with invertible 1x1 convolutions.
\newblock {\em Advances in neural information processing systems}, 31, 2018.

\bibitem{denseflow}
Matej Grci{\'c}, Ivan Grubi{\v{s}}i{\'c}, and Sini{\v{s}}a {\v{S}}egvi{\'c}.
\newblock Densely connected normalizing flows.
\newblock {\em Advances in Neural Information Processing Systems}, 34, 2021.

\bibitem{delta_vae}
Ali Razavi, A{\"{a}}ron van~den Oord, Ben Poole, and Oriol Vinyals.
\newblock Preventing posterior collapse with delta-vaes.
\newblock {\em CoRR}, abs/1901.03416, 2019.

\bibitem{spn}
Jacob Menick and Nal Kalchbrenner.
\newblock Generating high fidelity images with subscale pixel networks and
  multidimensional upscaling.
\newblock {\em arXiv preprint arXiv:1812.01608}, 2018.

\bibitem{ma2019macow}
Xuezhe Ma, Xiang Kong, Shanghang Zhang, and Eduard Hovy.
\newblock Macow: Masked convolutional generative flow.
\newblock {\em Advances in Neural Information Processing Systems}, 32, 2019.

\bibitem{locally}
Ajay Jain, Pieter Abbeel, and Deepak Pathak.
\newblock Locally masked convolution for autoregressive models.
\newblock In {\em Conference on Uncertainty in Artificial Intelligence}, pages
  1358--1367. PMLR, 2020.

\bibitem{image}
Niki Parmar, Ashish Vaswani, Jakob Uszkoreit, Lukasz Kaiser, Noam Shazeer,
  Alexander Ku, and Dustin Tran.
\newblock Image transformer.
\newblock In {\em International Conference on Machine Learning}, pages
  4055--4064. PMLR, 2018.

\bibitem{child2019generating}
Rewon Child, Scott Gray, Alec Radford, and Ilya Sutskever.
\newblock Generating long sequences with sparse transformers.
\newblock {\em arXiv preprint arXiv:1904.10509}, 2019.

\bibitem{udm_paper}
Dongjun Kim, Seungjae Shin, Kyungwoo Song, Wanmo Kang, and Il{-}Chul Moon.
\newblock Score matching model for unbounded data score.
\newblock {\em CoRR}, abs/2106.05527, 2021.

\bibitem{kingma2021variational}
Diederik~P Kingma, Tim Salimans, Ben Poole, and Jonathan Ho.
\newblock Variational diffusion models.
\newblock {\em arXiv preprint arXiv:2107.00630}, 2021.

\bibitem{sinha2021consistency}
Samarth Sinha and Adji~Bousso Dieng.
\newblock Consistency regularization for variational auto-encoders.
\newblock In A.~Beygelzimer, Y.~Dauphin, P.~Liang, and J.~Wortman Vaughan,
  editors, {\em Advances in Neural Information Processing Systems}, 2021.

\bibitem{chen2016variational}
Xi~Chen, Diederik~P Kingma, Tim Salimans, Yan Duan, Prafulla Dhariwal, John
  Schulman, Ilya Sutskever, and Pieter Abbeel.
\newblock Variational lossy autoencoder.
\newblock {\em arXiv preprint arXiv:1611.02731}, 2016.

\bibitem{Wang2004ImageQA}
Zhou Wang, A.~Bovik, H.~R. Sheikh, and E.~P. Simoncelli.
\newblock Image quality assessment: from error visibility to structural
  similarity.
\newblock {\em IEEE Transactions on Image Processing}, 13:600--612, 2004.

\bibitem{5bits_1}
Chris Finlay, Jörn-Henrik Jacobsen, Levon Nurbekyan, and Adam~M. Oberman.
\newblock How to train your neural ode: the world of jacobian and kinetic
  regularization.
\newblock In {\em ICML}, pages 3154--3164, 2020.

\bibitem{5bits_2}
Arash Vahdat, Karsten Kreis, and Jan Kautz.
\newblock Score-based generative modeling in latent space.
\newblock In A.~Beygelzimer, Y.~Dauphin, P.~Liang, and J.~Wortman Vaughan,
  editors, {\em Advances in Neural Information Processing Systems}, 2021.

\bibitem{banding_1}
Zhengzhong Tu, Jessie Lin, Yilin Wang, Balu Adsumilli, and Alan~C. Bovik.
\newblock Adaptive debanding filter.
\newblock {\em IEEE Signal Processing Letters}, 27:1715–1719, 2020.

\bibitem{banding_2}
Jatin Sapra, Zhou Wang, and Akshay Kapoor.
\newblock Capturing banding in images.
\newblock 08 2021.

\bibitem{rolinek2019variational}
Michal Rolinek, Dominik Zietlow, and Georg Martius.
\newblock Variational autoencoders pursue pca directions (by accident).
\newblock In {\em Proceedings of the IEEE/CVF Conference on Computer Vision and
  Pattern Recognition}, pages 12406--12415, 2019.

\bibitem{Duan2020Unsupervised}
Sunny Duan, Loic Matthey, Andre Saraiva, Nick Watters, Chris Burgess, Alexander
  Lerchner, and Irina Higgins.
\newblock Unsupervised model selection for variational disentangled
  representation learning.
\newblock In {\em International Conference on Learning Representations}, 2020.

\bibitem{impact1}
Elliott Gordon-Rodríguez, Gabriel Loaiza-Ganem, and John~P. Cunningham.
\newblock The continuous categorical: a novel simplex-valued exponential
  family.
\newblock In {\em ICML}, pages 3637--3647, 2020.

\bibitem{impact2}
Aaron Van Den~Oord, Oriol Vinyals, et~al.
\newblock Neural discrete representation learning.
\newblock {\em Advances in neural information processing systems}, 30, 2017.

\bibitem{impact3}
Rayhane Mama, Marc~S. Tyndel, Hashiam Kadhim, Cole Clifford, and Ragavan
  Thurairatnam.
\newblock {NWT:} towards natural audio-to-video generation with representation
  learning.
\newblock {\em CoRR}, abs/2106.04283, 2021.

\bibitem{impact4}
Alexei Baevski, Henry Zhou, Abdelrahman Mohamed, and Michael Auli.
\newblock wav2vec 2.0: {A} framework for self-supervised learning of speech
  representations.
\newblock {\em CoRR}, abs/2006.11477, 2020.

\bibitem{bottleneck_residual}
Kaiming He, Xiangyu Zhang, Shaoqing Ren, and Jian Sun.
\newblock Deep residual learning for image recognition.
\newblock In {\em 2016 IEEE Conference on Computer Vision and Pattern
  Recognition (CVPR)}, pages 770--778, 2016.

\bibitem{bergstra2012random}
James Bergstra and Yoshua Bengio.
\newblock Random search for hyper-parameter optimization.
\newblock {\em Journal of machine learning research}, 13(2), 2012.

\end{thebibliography}

\newpage
\appendix

\section{Additional architecture details}
\subsection{General model architecture}
We start by providing a more detailed information about the bidirectional HVAE architecture and how it works.

\begin{figure*}[t!]
    \centering
    \begin{subfigure}[t]{0.5\textwidth}
        \centering
        \includegraphics[width=\textwidth]{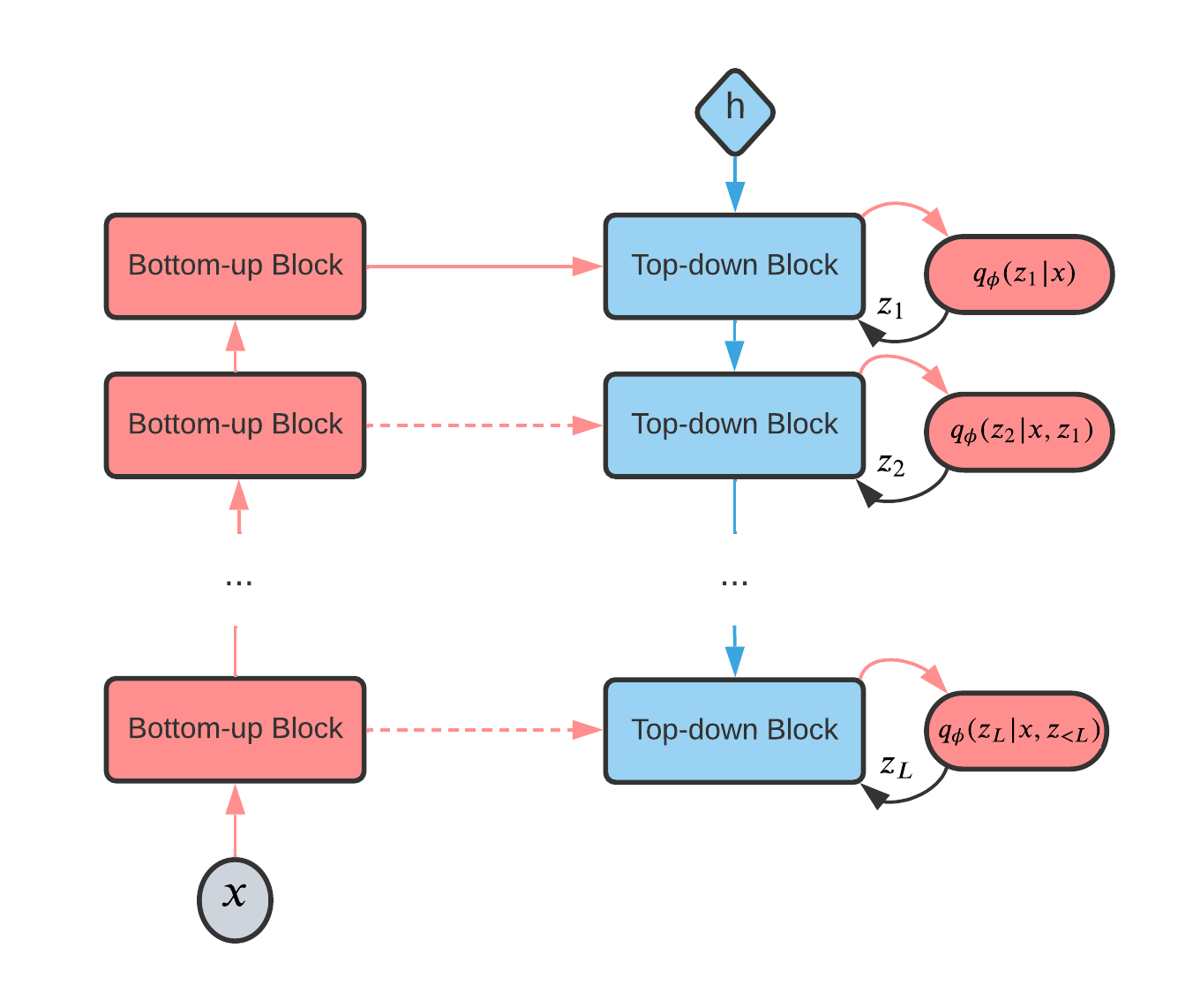}
        \caption{Inference model $q_\phi(z|x)$}
    \end{subfigure}%
    ~ 
    \begin{subfigure}[t]{0.5\textwidth}
        \centering
        \includegraphics[width=.58\textwidth]{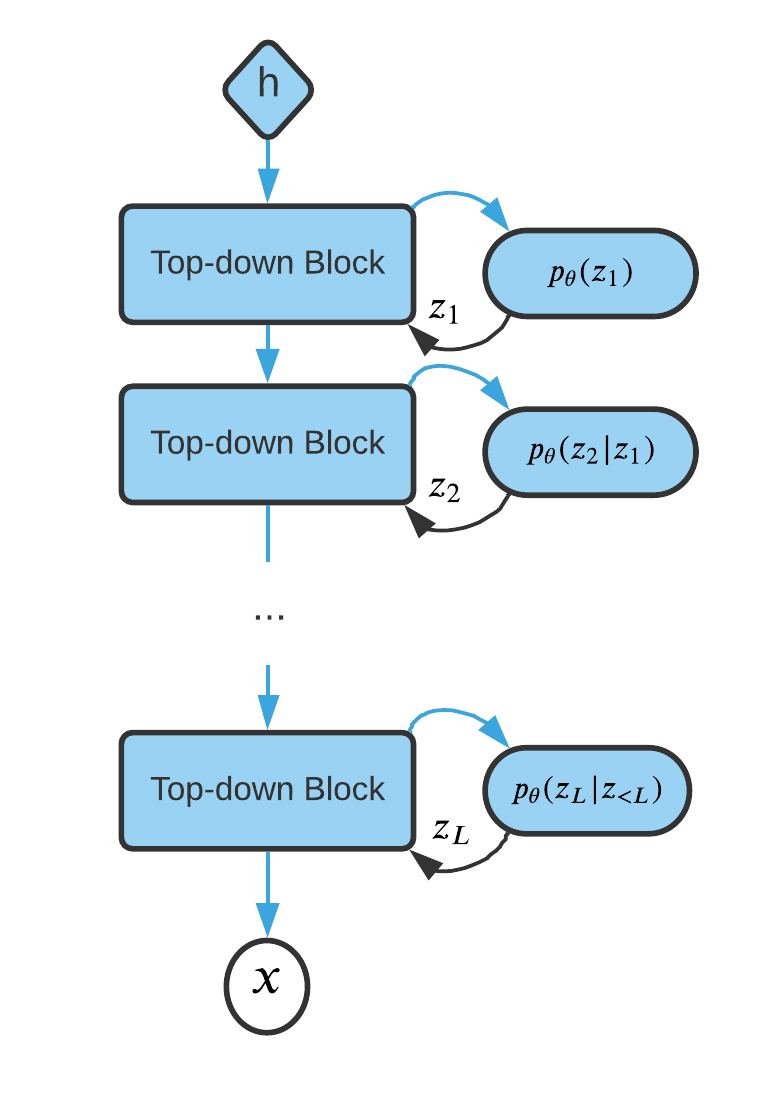}
        \caption{Generative model $p_\theta(x, z)$}
    \end{subfigure}
    \caption{\textbf{Bidirectional inference HVAE.} The blue blocks are shared between the inference model and generative model. The inference model's (encoder) outputs are in the bidirectional case computed by both a pass through the bottom-up then the top-down blocks. Dashed lines represent optional connections between the bottom-up and top-down blocks (more details in \ref{asymmetrical_design} and the \href{https://github.com/Rayhane-mamah/Efficient-VDVAE}{source code}).}
\label{bidirectional_hvae_fig}
\end{figure*}

Figure \ref{bidirectional_hvae_fig} depicts the general form of a bidirectional inference HVAE, color-coded with: red for the inference model and blue for the generative model. Shared parameters between the inference and generative model are also coded in blue. 

The bottom-up blocks in the HVAE extract latent activations $\chi$ that are used later on by the top-down block to create the posterior distribution $q_\phi(z_L|x, z_{<L})$. Both the prior and posterior distributions are created within the top-down block and are used in the KL divergence term of the ELBO.

During training, we sample $z_i$ from the posterior $q_\phi(z_i|x, z_{<i})$ which depends on the activations $\chi$ computed by the bottom-up blocks. During unconditional generation from the prior, the bottom-up blocks are unavailable, we thus sample $z_i$ from the prior $p_\theta(z_i|z_{<i})$ instead.

\begin{figure*}[t!]
    \centering
    \begin{subfigure}[t]{0.5\textwidth}
        \centering
        \includegraphics[width=.74\textwidth]{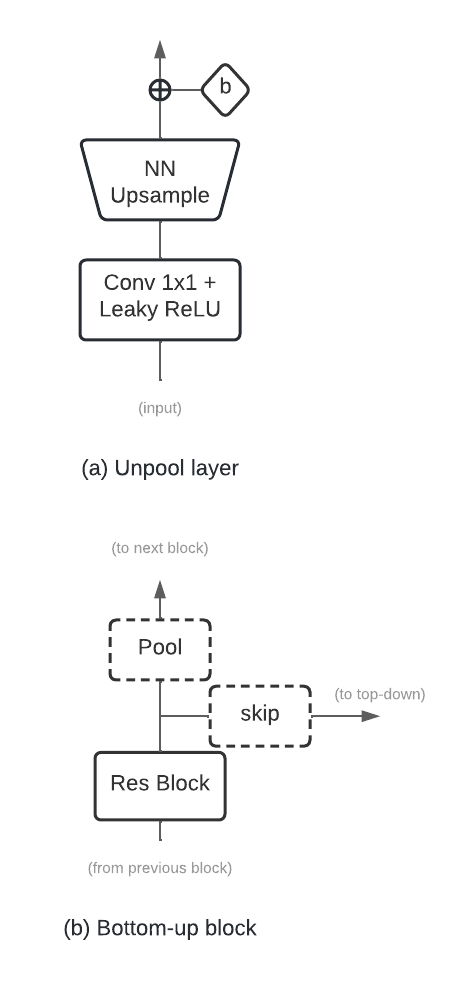}
    \end{subfigure}%
    ~ 
    \begin{subfigure}[t]{0.5\textwidth}
        \centering
        \includegraphics[width=\textwidth]{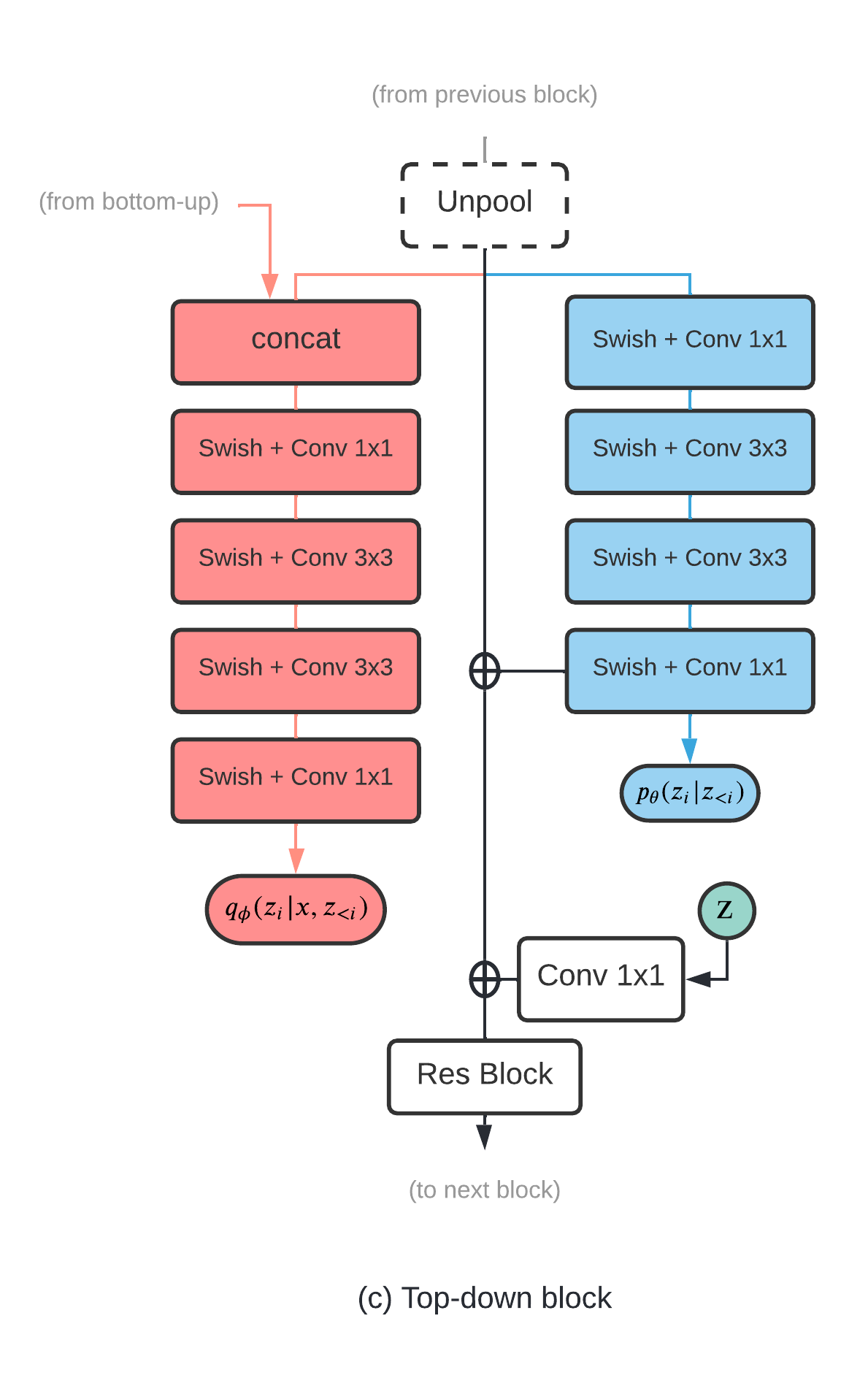}
    \end{subfigure}
    \caption{\textbf{Architecture of the Efficient-VDVAE.} Both the pooling and unpooling are different from the traditional VDVAE architecture for the sake of generalization of non-constant filter width in the model. The "Pool" layer is a convolution followed by a Leaky ReLU activation. Dashed blocks are optional and not always present (more details in the \href{https://github.com/Rayhane-mamah/Efficient-VDVAE}{source code}).}
\label{architecture_fig}
\end{figure*}

The VDVAE architecture is a special type of bidirectional HVAE, where the top-down block uses different bottleneck residual blocks\cite{bottleneck_residual} to compute the prior and posterior distributions. In figure \ref{architecture_fig}c, we show the color-coded top-down block from VDVAE.

In Efficient-VDVAE, we make three slight modifications to the VDVAE architecture:
\begin{itemize}
    \item \textbf{Bottom-up block:} We added a skip connection before propagating the output towards the top-down block. This enables us to project the activations $\chi$ to any arbitrary width when passing it to the posterior computation branch (in the top-down block), even if the filters number of the rest of the model is changing (figure \ref{architecture_fig}b).
    \item \textbf{Pool layer:} VDVAE uses a non-trainable average pooling to downsample activations. We replace that with a $1\times1$ convolution to have the freedom to change the number of filters.
    \item \textbf{Unpool layer:} We add a $1\times1$ convolution prior to the nearest neighbor upsampling to also have the freedom to change the filter size inside the top-down model (figure \ref{architecture_fig}a).
\end{itemize}

Although the VDVAE work raises the concern that using trainable convolutions in the "pool layer" and "unpool layer" can cause the low resolutions latent groups to not encode any information, we did not observe this behavior in our experiments. Nonetheless, should that happen on other datasets in the future, we provide the KL warm-up schedule described in NVAE\cite{vahdat2020nvae} as a solution in the \href{https://github.com/Rayhane-mamah/Efficient-VDVAE}{source code}. 

\subsection{Detailed layer distributions parameters}
In order to explain a part of the memory and NLL differences in table \ref{quantitative_compute}, we provide a more detailed overview of the distribution of the stochastic layers across different latent resolutions in table \ref{layer_dist}.

\begin{table}[]
\caption{Layers distribution by resolution. A single integer in the width column means that width was kept constant for the entire model (outside of the bottleneck residual blocks).}
\label{layer_dist}
\resizebox{\textwidth}{!}{%
\begin{tabular}{@{}ccccc@{}}
\toprule
\multicolumn{1}{c}{Dataset} & \multicolumn{1}{c}{Model} & Latent resolutions & Width & \multicolumn{1}{c}{Layers per resolution} \\ 
\midrule
\multirow{3}{*}{CIFAR-10} 
& C1 & $1^2, 4^2, 8^2, 16^2, 32^2$ & $384, 384, 192, 96, 48$ & $3, 4, 7, 11, 22$ \\ 
\cmidrule(){2-5} 
& C2 & $1^2, 4^2, 8^2, 16^2, 32^2$ & $384$ & $3, 4, 7, 11, 22$ \\ 
\cmidrule(){2-5} 
& VDVAE & $1^2, 4^2, 8^2, 16^2, 32^2$ & $384$ & $1, 3, 6, 11, 22$ \\ 
\midrule
\multirow{3}{*}{\begin{tabular}[c]{@{}c@{}}Imagenet \\ $32\times32$\end{tabular}} 
& C1 & $1^2, 4^2, 8^2, 16^2, 32^2$ & $512, 512, 256, 128, 64$ & $6, 7, 19, 25, 16$ \\ 
\cmidrule(){2-5} 
& C2 & $1^2, 4^2, 8^2, 16^2, 32^2$ & $512$ & $6, 7, 19, 25, 16$ \\ 
\cmidrule(){2-5}
& VDVAE & $1^2, 4^2, 8^2, 16^2, 32^2$ & $512$ & $2, 5, 10, 20, 41$ \\ 
\midrule
\multirow{3}{*}{\begin{tabular}[c]{@{}c@{}}Imagenet\\  $64\times64$\end{tabular}} 
& C1 & $1^2, 4^2, 8^2, 16^2, 32^2, 64^2$ & $512, 512, 256, 256, 64, 64$ & $6, 7, 19, 25, 16, 11$ \\ 
\cmidrule(){2-5} 
& C2 & $1^2, 4^2, 8^2, 16^2, 32^2, 64^2$ & $512$ & $6, 7, 19, 25, 16, 11$ \\ 
\cmidrule(){2-5}
& VDVAE & $1^2, 4^2, 8^2, 16^2, 32^2, 64^2$ & $512$ & $2, 4, 8, 16, 32, 13$ \\ 
\midrule
\multirow{3}{*}{\begin{tabular}[c]{@{}c@{}}FFHQ\\  $256\times256$\end{tabular}}   
& C1 & $1^2, 4^2, 8^2, 16^2, 32^2, 64^2, 128^2, 256^2$ & $512, 512, 512, 256, 256, 128, 128, 128$ & $2, 4, 5, 10, 22, 14, 8, 1$ \\
\cmidrule(){2-5} 
& C2 & $1^2, 4^2, 8^2, 16^2, 32^2, 64^2, 128^2, 256^2$ & $512$ & $2, 4, 5, 10, 22, 14, 8, 1$ \\
\cmidrule(){2-5}
& VDVAE & $1^2, 4^2, 8^2, 16^2, 32^2, 64^2, 128^2, 256^2$ & $512$ & $2, 4, 5, 10, 22, 14, 8, 1$ \\                             
\bottomrule
\end{tabular}
}
\end{table}

\section{Additional Results}
\subsection{Gradient smoothing and divergence}\label{extreme_gradient_smoothing}
All the models in this work, including the CelebAHQ $1024\times1024$ and FFHQ $1024\times1024$ models, were trained with a maximal memory constraint of $320$GB. As such, the maximum batch size that we were able to use on high resolution datasets is $8$, which amplified the instabilities of not using gradient smoothing. As seen in table \ref{extreme_gradient_smoothing_table}, high resolution models cannot be trained without gradient smoothing as they very quickly diverge, despite our attempts of reducing the learning rate to account for the small batch size.

\subsection{Gradient smoothing and model depth}\label{depth_stability}
For completeness, we train networks while fixing all hyper-parameters and only changing the depth of the model. We then re-train the networks while removing the gradient smoothing and measure both NLL and the number of skipped updates. We show our results in table \ref{depth_stability_table}.

\begin{table}[]
\vskip 0.1in
\centering
\begin{tabular}{@{}ccccccc@{}}
\toprule
\multirow{2}{*}{Model depth} &
   & \multicolumn{2}{c}{Without gradient smoothing} &
   & \multicolumn{2}{c}{With gradient smoothing} \\ 
\cmidrule(){3-4} \cmidrule(){6-7} 
 && NLL (bits/dim) & Skipped updates && NLL (bits/dim) & Skipped updates \\ 
\midrule
$30$ && $2.28$ & $2$ && $2.27$ & $0$ \\ 
$49$ && $2.20$ & $9$ && $2.21$ & $1$ \\ 
$66$ && $2.18$ & $14$ && $2.18$ & $0$ \\ 
$69$ && $2.18$ & $24$ && $2.17$ & $0$ \\ 
$75$ && $2.17$ & $37$ && $2.17$ & $2$ \\ 

\bottomrule
\end{tabular}
\caption{Effect of gradient smoothing with respect to model depth on 8-bit FFHQ $256\times256$ after $800$k updates with $\beta=\log(2)$. The gradient smoothing doesn't affect NLL, but greatly reduces the number of skips as the model gets deeper (All models were run with a batch size of $8$).}
\label{depth_stability_table}
\end{table}

\section{Additional tips and insights for practitioners}\label{tips_insights}
In this section, we provide some notable tips and insights from our experience working with Efficient-VDVAE in the hope of saving valuable time for practitioners who want to use similar architectures.

\subsection{Taming the exponentially growing variance of deep models}
in their work\cite{child2021very}, VDVAE's author demonstrates how scaling the last convolution of each bottleneck residual block by $1/\sqrt{L}$, $L$ being the total number of latent hierarchical variables, helps stabilize the model's training. In their open sourced codebase, they additionally scale the weights of the projection of sample $z$ (figure \ref{architecture_fig}c).

When training under the Efficient-VDVAE setup, we observe that scaling the $z$ projections is much more important than scaling the last convolution of each residual block. Not doing the former results in models diverging at the first training step, while not doing the latter can still result in models that train until convergence.

\subsection{Asymmetrical model design}\label{asymmetrical_design}

Traditionally, HVAEs were built to have symmetrical bottom-up and top-down models, where each top-down block $i$ receives a different activation $\chi_i$ from a bottom-up block $L - i$. VDVAE breaks this symmetry by allowing $K$ consecutive top-down blocks $\{i, i+1, ..., i+K\}$ to use the same activation $\chi_i$. More specifically, $\chi_i$ is only extracted from the bottom-up model once per latent resolution, and is used for all top-down blocks of that resolution.

In our \href{https://github.com/Rayhane-mamah/Efficient-VDVAE}{source code}, we view these two designs as opposite extreme ends of the same general spectrum. In fact, we implement our models so that Efficient-VDVAE can be designed to be symmetrical, asymmetrical and any combination of the two (symmetrical in parts). 

Empirically, for two models with the same number of layers, we didn't notice any advantage to either design on NLL, convergence or stability. However, we experimented with reducing the bottom-up model's size in an attempt to reduce computational requirements without hurting the model's performance. We report our findings in table \ref{asymmetrical_table}. We noticed that it is in fact possible to reduce the bottom-up model's complexity up to a certain limit before starting to lose performance, as measured in NLL.

\begin{table}[!]
\centering
\begin{tabular}{@{}cclcc@{}}
\toprule
\multicolumn{2}{c}{Network depth} &
   &
  \multirow{2}{*}{\begin{tabular}[c]{@{}c@{}}Memory \\ (GB)\end{tabular}} &
  \multirow{2}{*}{\begin{tabular}[c]{@{}c@{}}NLL\\ (bits/dim)\end{tabular}} \\ \cmidrule(){1-2}
Bottom-up & Top-down &&       &       \\ 
\midrule
$22$ & $47$ && $13$ & $2.90$ \\
$27$ & $47$ && $15$ & $2.89$ \\
$33$ & $47$ && $17$ & $2.87$ \\
$40$ & $47$ && $20$ & $2.87$ \\
$47$ & $47$ && $22$ & $2.87$ \\
\bottomrule
\end{tabular}
\caption{Effect of reducing the bottom-up model's depth on CIFAR-10.}
\label{asymmetrical_table}
\end{table}

\subsection{A note on the resolution of the latent space}
We want to allocate this section to talk about the effect of building latent spaces in the same resolution of the input image. 

While it is common practice in autoencoders to have a latent space smaller than the image so that the network compresses the pixel information and avoid learning an identity function, VAEs practically never learn an optimal latent space to model a distribution $p(x)$, even if they have the capacity to do so\cite{yu2020frombayestoit}.

The implementation of latent spaces in the same resolution as the image can boost the NLL of the HVAE as shown in table \ref{high_res_nll}. Qualitatively, however, these layers barely have a visible effect as they mostly operate on the noise of the pixels. Thus, if the goal is generative modeling, then the high resolution layers can be dropped for a memory load gain without major side effects.

\begin{table}
\centering
\begin{tabular}{@{}ccccclcc@{}}
\toprule
\multicolumn{5}{c}{Distribution of Layers} &
   &
  \multirow{2}{*}{\begin{tabular}[c]{@{}c@{}}Memory \\ (GB)\end{tabular}} &
  \multirow{2}{*}{\begin{tabular}[c]{@{}c@{}}NLL\\ (bits/dim)\end{tabular}} \\ \cmidrule(){1-5}
$32 \times 32$ & $16 \times 16$ & $8 \times 8$ & $4 \times 4$ & $1 \times 1$ &  &       &       \\ 
\midrule
$0$ & $11$ & $7$ & $4$ & $3$ && $9$ & $2.96$ \\
$11$ & $11$ & $7$ & $4$ & $3$ && $13$ & $2.91$ \\
$22$ & $11$ & $7$ & $4$ & $3$ && $17$ & $2.87$ \\
\bottomrule
\end{tabular}
\caption{Effect of using high resolution layers on CIFAR-10. These results hold for all other datasets we tried except for MNIST because it has a binary distribution that we model using the Bernoulli distribution.}
\label{high_res_nll}
\end{table}

\subsection{A note on the overfitting of HVAEs}

Like any other maximum likelihood model, HVAEs will overfit when they can. HVAE's overfitting usually manifests as a large discrepancy between the aggregate posterior of the training data and the aggregate posterior of the validation data. This discrepancy can be measured by a difference between $D_\KL(q_\phi(z|x) || p_\theta(z))$ in training and validation.

CIFAR-10 is a good example of when HVAEs (and maximum likelihood models in general) overfit, which is the main reason why models that rely on augmentations score better on that benchmark. CR-NVAE\cite{sinha2021consistency} is a good example of an attempt to reduce the HVAE's overfitting on CIFAR-10 by applying a consistency regularization on the latent space. While CR-NVAE was built on top of NVAE in their work, there should be no problem with applying the same consistency regularization to any other HVAE. That is however outside the scope of this work.

\subsection{Keep a situational mindset}

Despite the theory and the rules of thumb, it is always healthy to keep a situational mindset. At the end of the day, "For most data sets only a few of the hyper-parameters really matter, but [...] different hyper-parameters are important on different data sets"\cite{bergstra2012random}.

\section{Additional samples}\label{additional_samples_appendix}
\subsection{Pruned posteriors reconstructions}
\begin{figure}
    \centering
    \includegraphics[width=\textwidth]{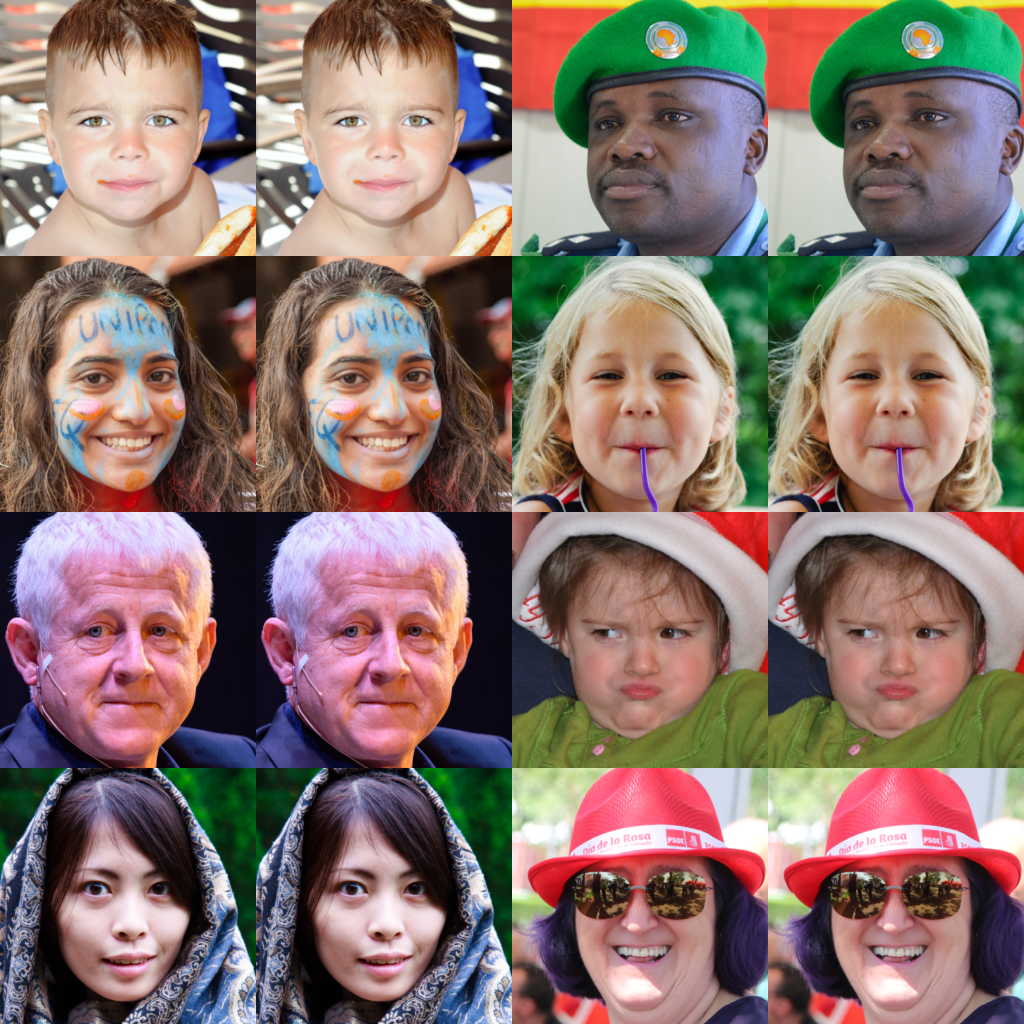}
    \caption{Image reconstructions with $100\%$ of the posteriors (left) and $3\%$ of the posteriors with the highest average KL divergence (right). Reconstruction differences are indistinguishable.}
    \label{3percent_reconstructions_fig}
\end{figure}

We show in figure \ref{3percent_reconstructions_fig} reconstructions from Efficient-VDVAE before and after pruning $97\%$ of its posteriors. Qualitative observations agree with the reconstruction loss, and no noticeable differences appear on the images. We hypothesize from these observations that the $3\%$ of posteriors used in these reconstructions encode most of the latent information and they can be sufficient, and memory efficient, to use in downstream tasks.

\FloatBarrier
\subsection{Generated samples from the prior}\label{additional_prior_samples_appendix}
Due to lack of space in the paper, we share additional samples generated from the prior distribution here.

\begin{figure}[!htb]
    \centering
    \includegraphics[width=.5\textwidth]{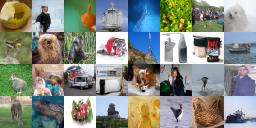}
    \caption{More unconditional samples on Imagenet $32\times32$ ($t=0.85$).}
    \label{more_imagenet32_fig}
\end{figure}

\begin{figure}[!htb]
    \centering
    \includegraphics[width=.6\textwidth]{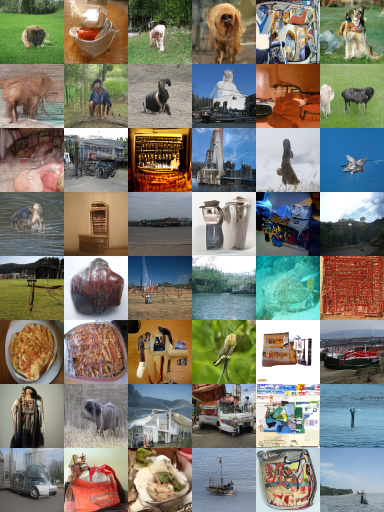}
    \caption{More unconditional samples on Imagenet $64\times64$ ($t=0.9$).}
    \label{more_imagenet64_fig}
\end{figure}

\begin{figure}[!htb]
    \centering
    \includegraphics[width=.5\textwidth]{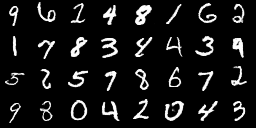}
    \caption{Unconditional samples on MNIST ($t=1$).}
    \label{more_mnist_fig}
\end{figure}

\begin{figure}[!htb]
    \centering
    \includegraphics[width=.6\textwidth]{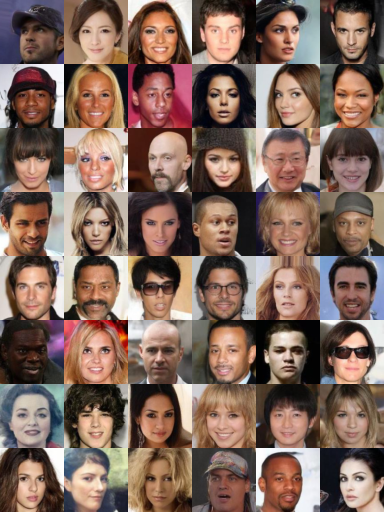}
    \caption{More unconditional samples on CelebA $64\times64$ ($t=0.8$).}
    \label{more_celeba64_fig}
\end{figure}

\begin{figure}[!htb]
    \centering
    \includegraphics[width=.85\textwidth]{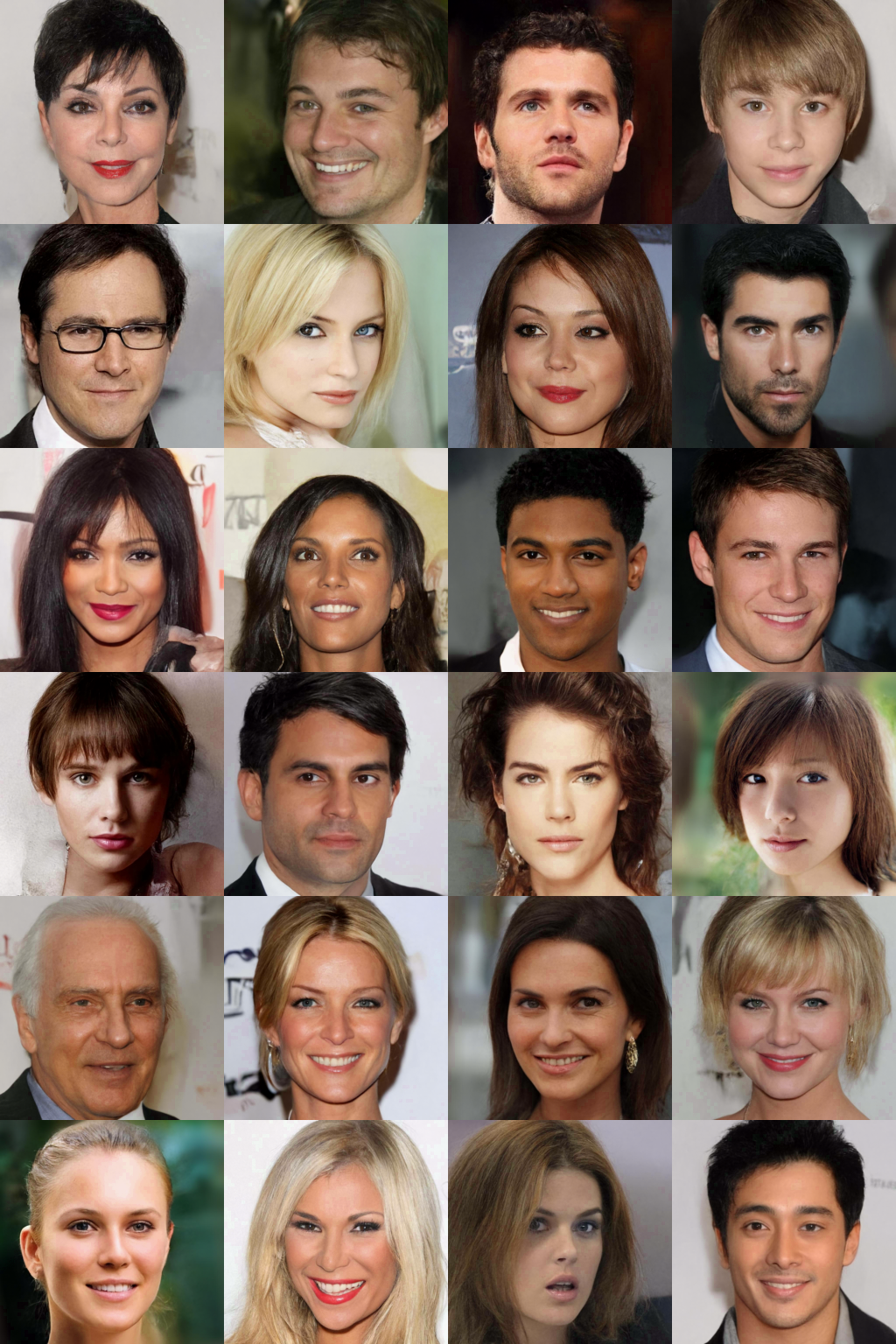}
    \caption{More unconditional samples on 5-bit CelebAHQ $256\times256$ ($t=0.85$).}
    \label{more_celebahq256_5bits_fig}
\end{figure}

\begin{figure}[!htb]
    \centering
    \includegraphics[width=.85\textwidth]{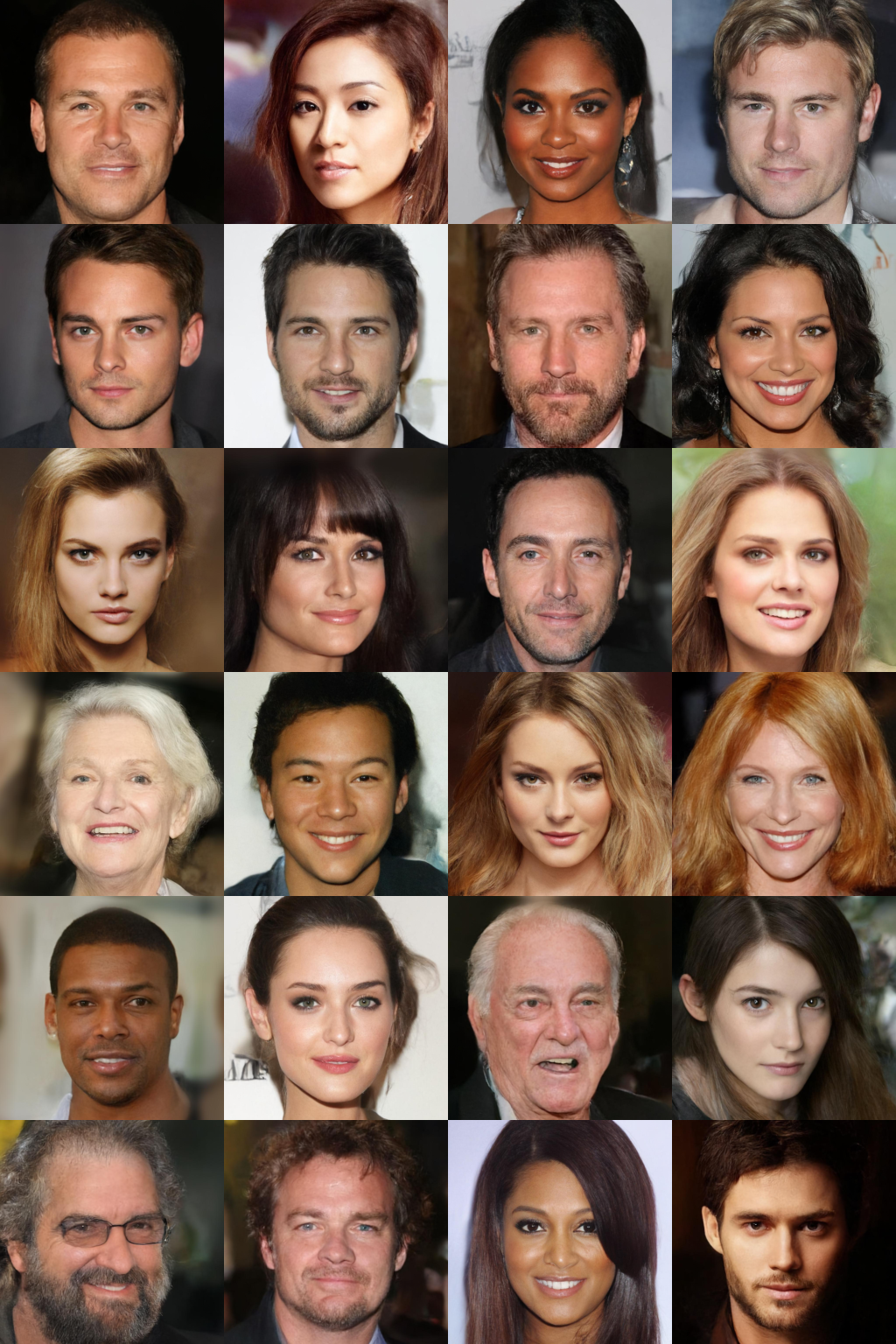}
    \caption{More unconditional samples on 8-bit CelebAHQ $256\times256$ ($t=0.85$).}
    \label{more_celebahq256_8bits_fig}
\end{figure}

\begin{figure}[!htb]
    \centering
    \includegraphics[width=.85\textwidth]{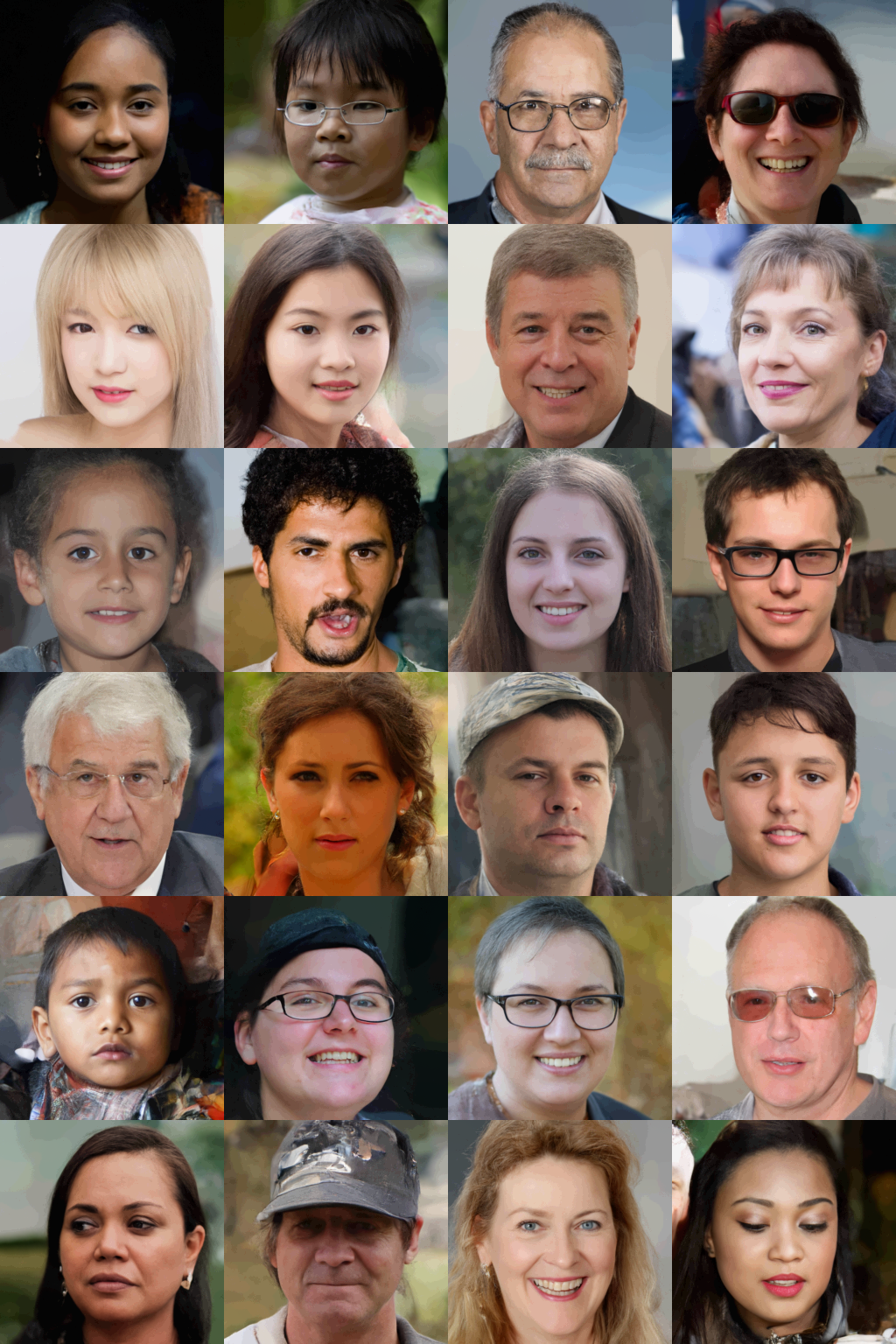}
    \caption{More unconditional samples on 5-bit FFHQ $256\times256$ ($t=0.8$).}
    \label{more_ffhq256_5bits_fig}
\end{figure}

\begin{figure}[!htb]
    \centering
    \includegraphics[width=.85\textwidth]{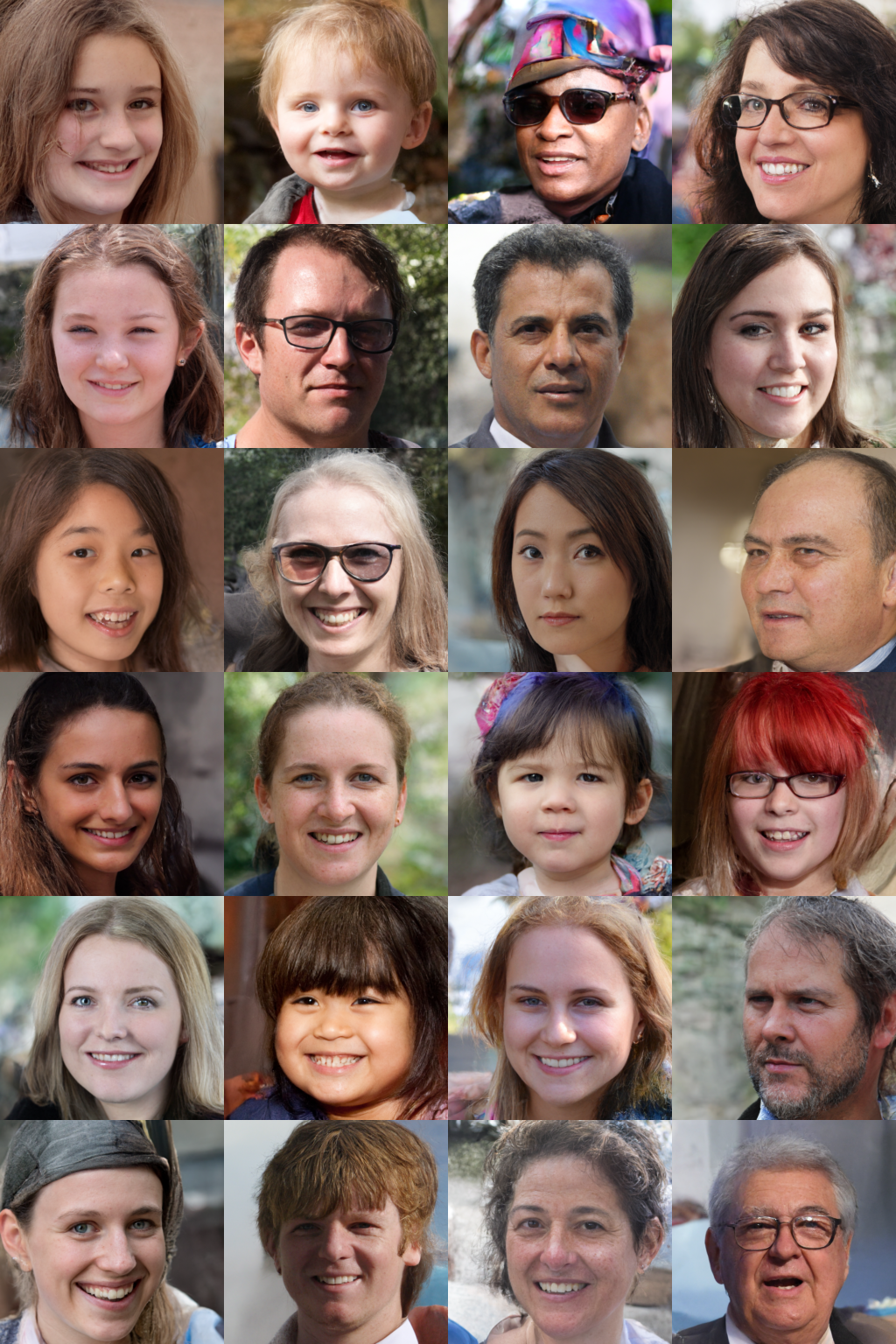}
    \caption{More unconditional samples on 8-bit FFHQ $256\times256$ ($t=0.8$).}
    \label{more_ffhq256_8bits_fig}
\end{figure}

\begin{figure}[!htb]
    \centering
    \includegraphics[width=.85\textwidth]{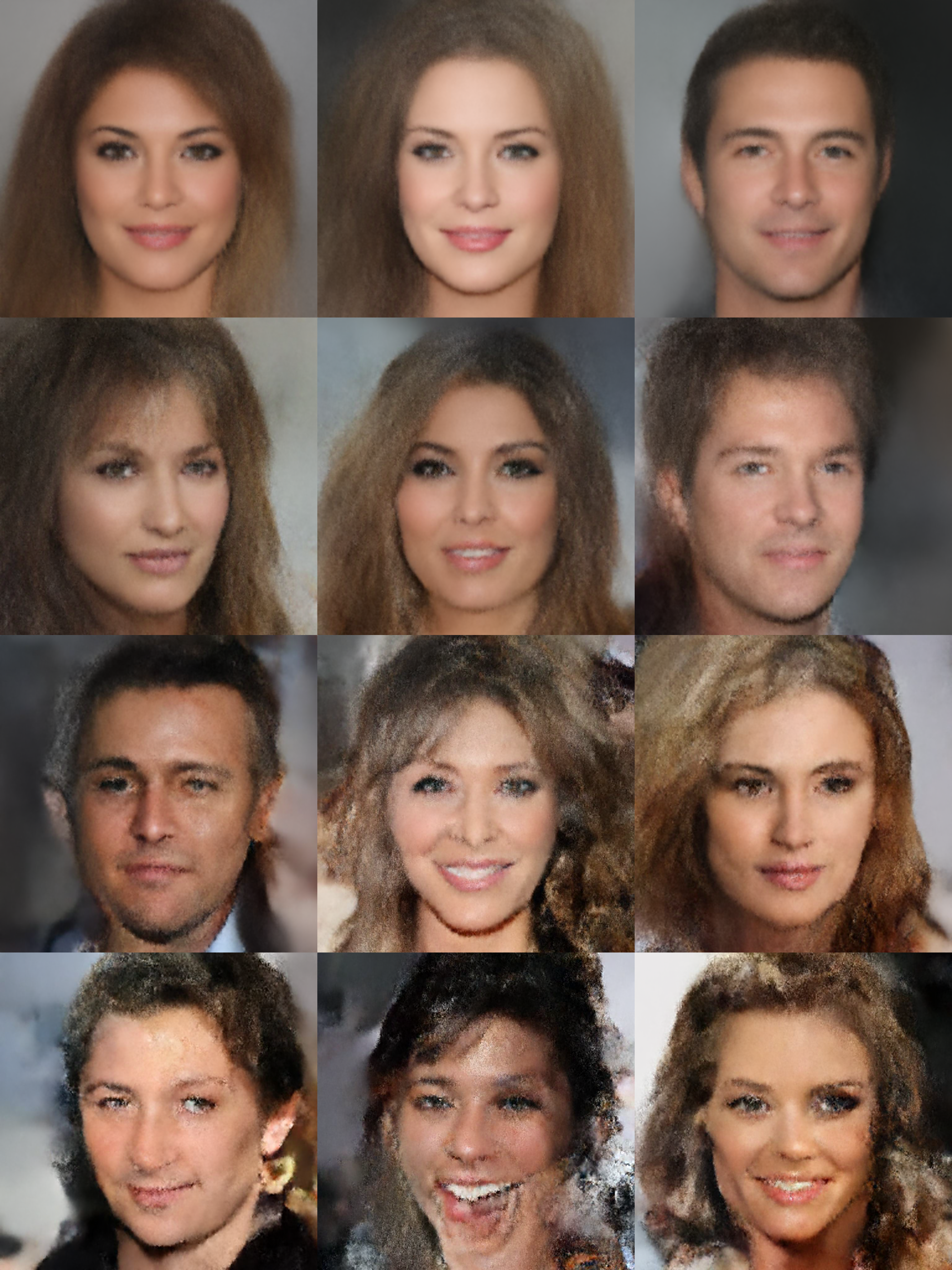}
    \caption{More unconditional samples on CelebAHQ $1024\times1024$. From top row to bottom row, temperatures are $\{0.4, 0.6, 0.8, 1.\}$}
    \label{more_celebahq1024_fig}
\end{figure}

\begin{figure}[!htb]
    \centering
    \includegraphics[width=.85\textwidth]{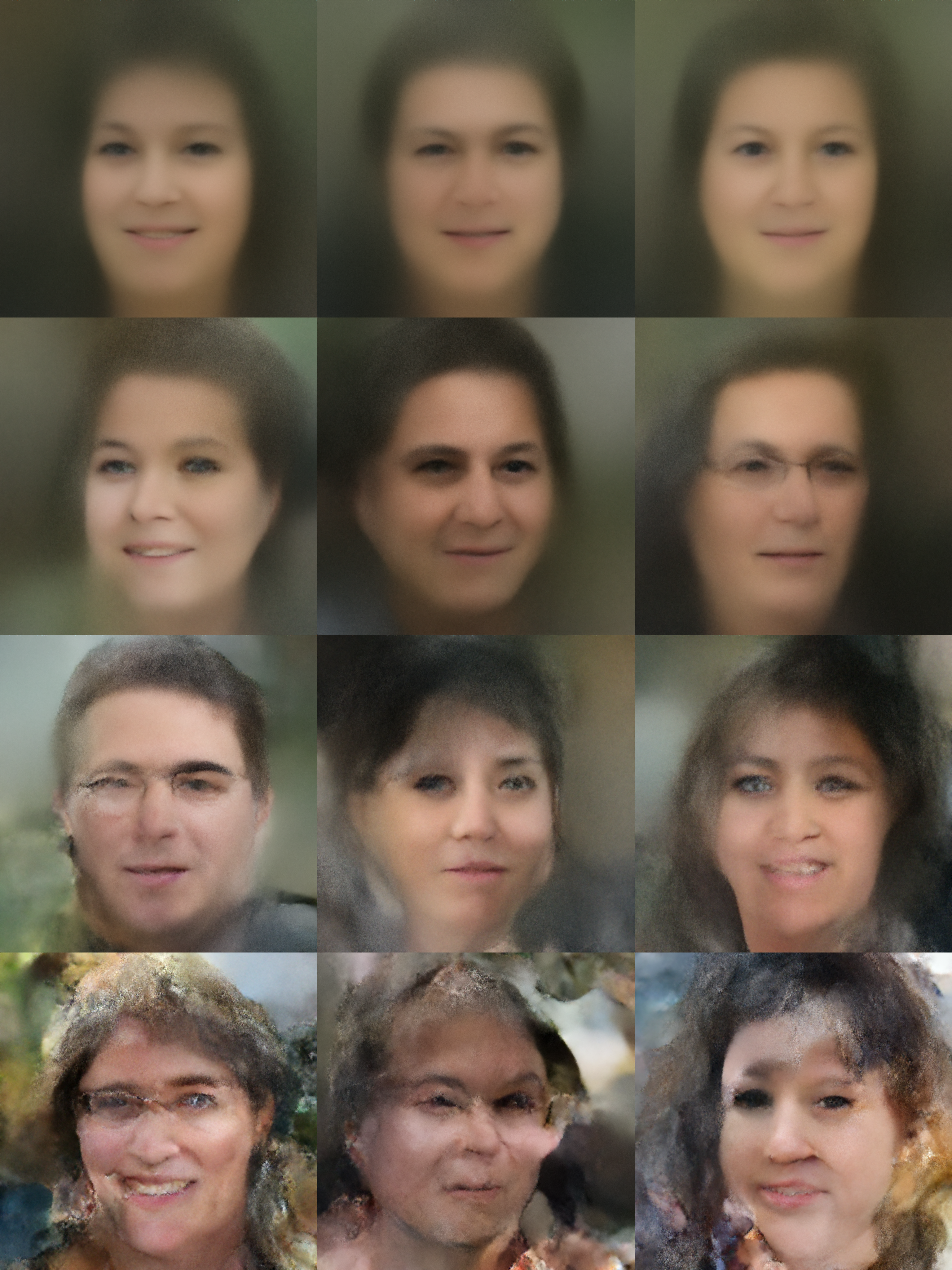}
    \caption{More unconditional samples on FFHQ $1024\times1024$. From top row to bottom row, temperatures are $\{0.3, 0.5, 0.7, 0.9\}$}
    \label{more_ffhq1024_fig}
\end{figure}

\end{document}